\documentclass[10pt,twocolumn,letterpaper]{article}

\usepackage{cvpr}              

\usepackage{graphicx}
\usepackage{amsmath}
\usepackage{amssymb}
\usepackage{booktabs}
\usepackage[pagebackref,breaklinks,colorlinks]{hyperref}
\usepackage{url}
\usepackage{subcaption}

\usepackage{multirow}
\usepackage{color,soul}
\usepackage[dvipsnames]{xcolor}
\usepackage{esvect}
\usepackage{xspace}
\usepackage{array}
\usepackage{subcaption}
\usepackage[accsupp]{axessibility}  
\usepackage{hyperref}

\usepackage[capitalize]{cleveref}
\crefname{section}{Sec.}{Se
cs.}
\Crefname{section}{Section}{Sections}
\Crefname{table}{Table}{Tables}
\crefname{table}{Tab.}{Tabs.}

\newcommand{\nickname}{CLIP-Sculptor\xspace}

\def\cvprPaperID{12333} 
\def\confName{CVPR}
\def\confYear{2023}

\begin{document}

\title{\nickname: Zero-Shot Generation of High-Fidelity and Diverse Shapes from Natural Language}

\author{Aditya Sanghi\footnotemark[2] \qquad Rao Fu\footnotemark[3] \qquad Vivian Liu\footnotemark[4] \qquad Karl D.D. Willis\footnotemark[2] \qquad Hooman Shayani\footnotemark[2] \\Amir H. Khasahmadi\footnotemark[2] \qquad Srinath Sridhar\footnotemark[3] \qquad Daniel Ritchie\footnotemark[3] \\
\text{\normalsize Autodesk Research\footnotemark[2] \qquad Brown University\footnotemark[3] \qquad Columbia University\footnotemark[4]}\\
{\color{magenta} \url{https://ivl.cs.brown.edu/\#/projects/clip-sculptor}}
}

\twocolumn[{%
\renewcommand\twocolumn[1][]{#1}%
\maketitle

\begin{center}
\centering
\captionsetup{type=figure}
\setlength{\tabcolsep}{0pt}
        \includegraphics[width=\textwidth]{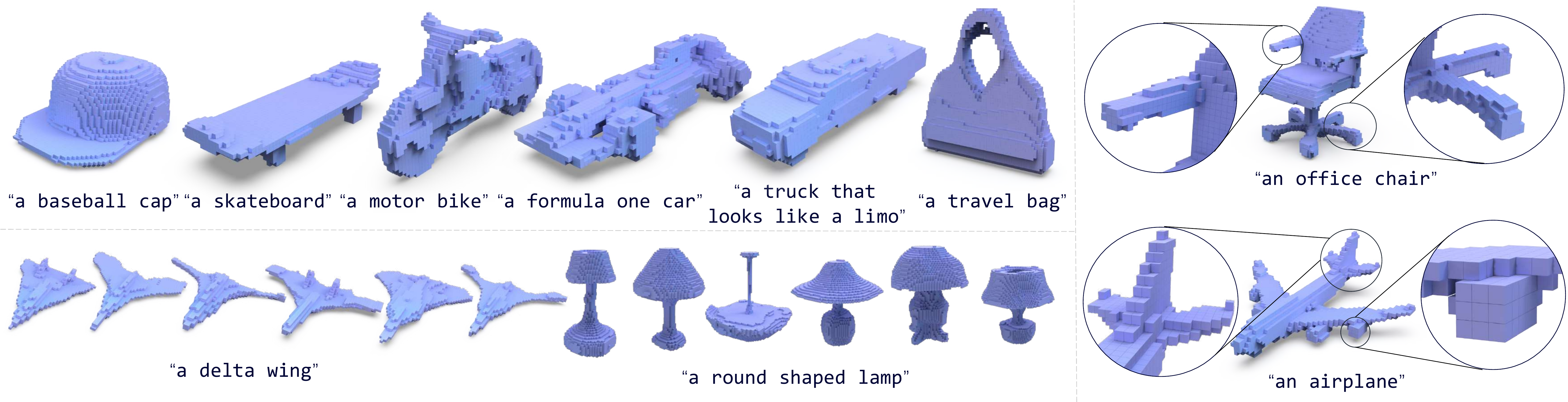}
        \caption{\nickname is a zero-shot text-to-shape generation method. Left Top/Bottom: It generates \textbf{diverse} shapes that reflect the semantic meaning of the text input \textbf{without requiring any text data} during training. Right: The method also generates  \textbf{high-fidelity} shapes.
        }
        \label{fig:teaser}
        \vspace{-0.05in}
\end{center}%
}]


\begin{abstract}

Recent works have  demonstrated that natural language can be used to generate and edit 3D shapes. However, these methods generate  shapes with limited fidelity and diversity.
We introduce \nickname, a method to address these constraints by producing high-fidelity and diverse 3D shapes
without the need for (text, shape) pairs during training.
\nickname achieves this in a multi-resolution approach that first generates in a low-dimensional latent space and then upscales to a higher resolution for improved shape fidelity.
For improved shape diversity, we use a discrete latent space which is modeled using a transformer conditioned on CLIP's image-text embedding space. We also present a novel variant of classifier-free guidance, which improves the accuracy-diversity trade-off. Finally, we perform extensive experiments demonstrating that \nickname outperforms state-of-the-art baselines.

\end{abstract}

\section{Introduction}

In recent years, there has been rapid progress in text-conditioned image generation~\cite{ramesh2021zero,ramesh2022hierarchical,Rombach_2022_CVPR}, which has been driven by advances in multimodal understanding learned from web-scaled paired (text, image) data.
These advances have led to applications in domains ranging from content creation~\cite{ramesh2022hierarchical, 3dalle, liu2022opal} to human--robot interaction~\cite{shridhar2021cliport}.
Unfortunately, developing the analogue of a text-conditioned 3D shape generator is challenging because it is difficult to obtain (text, 3D shape) pairs at large scale. Prior work has attempted to address this problem by collecting text-shape paired data ~\cite{chen2018text2shape,liu2022towards,mittal2022autosdf,shapecrafter,shapeglot}, but these approaches have been limited to a small number of object categories.


A promising way around this data bottleneck is to use weak supervision from large-scale vision/language models such as CLIP~\cite{radford2021learning}.
One approach is to directly optimize a 3D representation such that (text, image render) pairs are 
aligned when projected into the CLIP embedding space. Prior work has applied this approach to stylize 3D meshes~\cite{michel2022text2mesh,wang2022clip} and to create abstract ``dreamlike'' objects using neural radiance fields ~\cite{jain2022zero} or meshes  ~\cite{khalid2022clip}.
However, neither of the aforementioned methods produce realistic object geometry, and they can  require expensive optimization.
Another approach, more in line with text-to-image generators~\cite{ramesh2021zero,ramesh2022hierarchical}, is to train a 
conditional generative model.
The CLIP-Forge system~\cite{sanghi2022clip} builds such a model without paired (text, shape) data by using rendered images of shapes at training time and leveraging the CLIP embedding space to bridge the 
modalities of image and text at inference time. 
CLIP-Forge demonstrates compelling zero-shot generation abilities but produces low-fidelity shapes which do not capture the full diversity of shapes found in the training distribution.

In this paper, we propose \nickname, a text-conditioned 3D shape generative model that outperforms the state of the art by improving shape diversity and fidelity with only (image, shape) pairs as supervision.
\nickname's novelty lies in its multi-resolution, voxel-based conditional generation scheme. Without (text, shape) pairs, \nickname learns to generate 3D shapes of common object categories from the text-image joint embedding of CLIP.
To achieve high fidelity outputs, \nickname adopts a multi-resolution approach: it first generates a low-resolution latent grid representation that captures the semantics from text/image, then upscales it to a higher resolution latent grid representation with a super-resolution model, and finally decodes the output geometry.
To generate diverse shapes, \nickname adopts discrete latent representations which are obtained using a vector quantization scheme that avoids posterior collapse.
To further improve shape fidelity and diversity, \nickname uses a masked transformer architecture. We additionally propose a novel annealed strategy for the classifier-free guidance~\cite{ho2022classifier}.
To sum up, we make the following contributions:
\begin{itemize}
    \item \nickname, a multi-resolution text-conditional shape generative model that achieves both high fidelity and diversity without the need for (text, shape) pairs.
    \item A novel variant of classifier-free guidance for generative models, with an annealed guidance schedule to achieve better quality for a given diversity level.
\end{itemize}

\begin{table}

\centering
\caption{High-level comparison between zero-shot text-to-shape generation methods. Inference time is calculated using a single NVIDIA Tesla V100 GPU. }
\label{tab:intro_table}
 \begin{tabular}{c|r|c|c}
\toprule
\textit{Method} & \textbf{Inference} & \textbf{Fidelity} &   \textbf{Diversity}  \\
\midrule
DreamFields \cite{jain2022zero} &  \color{Red} $>$ 24 hrs~~~ & \color{Red}  Low & \color{Red} Single\\
Clip-Mesh \cite{khalid2022clip}&  \color{Red} 30~min ~~~& \color{Red}  Low & \color{Red}  Single  \\
Clip-Forge \cite{sanghi2022clip}& \color{Green} 6.41 ms~~~ & \color{Orange} Medium & \color{Orange} Medium  \\
\nickname & \color{Orange}0.91 sec~~~ & \color{Green} High & \color{Green} High\\
\bottomrule

\end{tabular}
\end{table}
\section{Related Work}

\noindent \textbf{Neural Discrete Representation.}
Discrete representations \cite{oord2017neural} were first proposed for image generation in the context of variational autoencoders as a method to improve image quality and avoid posterior collapse. 
Further work introduced multi-scale and hierarchical variants \cite{razavi2019generating, dhariwal2020jukebox} which improved generative capabilities. Recently, 3D discrete representations have emerged as well, such as discretized voxel and implicit grids \cite{yan2022shapeformer,mittal2022autosdf}. In this work, we take inspiration from hierarchical VQ-VAEs  \cite{razavi2019generating, dhariwal2020jukebox} and propose an architecture of hierarchical discrete representations capable of generating high fidelity 3D shapes.

\noindent \textbf{Latent Generative Models.}
In recent years, latent generative models have been widely adopted, because these models can effectively generate low-dimensional latent representations which can  be used to  generate high fidelity images and shapes. In the 3D domain,  GAN-based latent models   \cite{achlioptas2018learning, chen2019learning,ibing20213d} have shown impressive results but tend to suffer from training instability and mode collapse. Flow-based models \cite{yang2019pointflow,sanghi2022clip} have been proposed to alleviate these problems, but they yield sample quality that is inferior to GAN-based models.
Recent works in the image domain use diffusion \cite{gu2022vector, rombach2022high} or masking models \cite{chang2022maskgit} on the latent space which can increase the inference efficiency and still give quality outputs. Building on these works, we propose a hierarchical latent generative model that can further improve the fidelity and diversity of shape generations. 

\noindent \textbf{Classifier-free Guidance.}
Classifier-free guidance, proposed by \cite{ho2022classifier}, jointly trains a conditional and unconditional generative model, whose score estimates are mixed to establish a trade-off between diversity measured by Frechet inception distance (FID) and sample accuracy/fidelity measured by Inception score (IS). Classifier-free guidance guides unconditional samples in the direction of conditional ones and has been implemented in practice to achieve state of the art results in recent work \cite{rombach_highres, makeascene, gal22image, gu2022vector}. To the best of our knowledge, guidance scale has always been kept constant outside the concurrent exploration in  \cite{imagenvideo}. Their oscillating guidance technique, however does not show improvements in sample fidelity and produces more artifacts in their generations. In this work, we find that rather using annealed scheduling can give better quality versus diversity trade-offs.

\noindent \textbf{Text-to-Shape Generation.}
Text-to-shape generation has gained momentum in recent years. Recent works\cite{chen2018text2shape,liu2022towards,mittal2022autosdf,shapecrafter} use supervised text-shape pairs to generate shapes effectively using natural language. However, a major drawback is the lack of available text-shape datasets, which limits supervised methods to  generate shapes only in  few categories. To solve this, several recent works have successfully leveraged the image-text embeddings of CLIP as a prior \cite{radford2021learning} by converting shapes into images using renders. One line of work uses differentiable renderers \cite{michel2022text2mesh,jain2022zero, poole2022dreamfusion, khalid2022clip, wang2022clip}. Another line of work learns a mapping from the image space to the shape space \cite{sanghi2022clip}. Although these methods are effective, they suffer from either long optimization time \cite{ jain2022zero, poole2022dreamfusion, khalid2022clip, michel2022text2mesh} or limited quality of generated shapes \cite{sanghi2022clip, jain2022zero, khalid2022clip}. Our method is able to generate more diverse shapes of higher fidelity within a comparably short inference time.

\begin{figure*}[t!]
    \begin{center}
        \includegraphics[width=\textwidth]{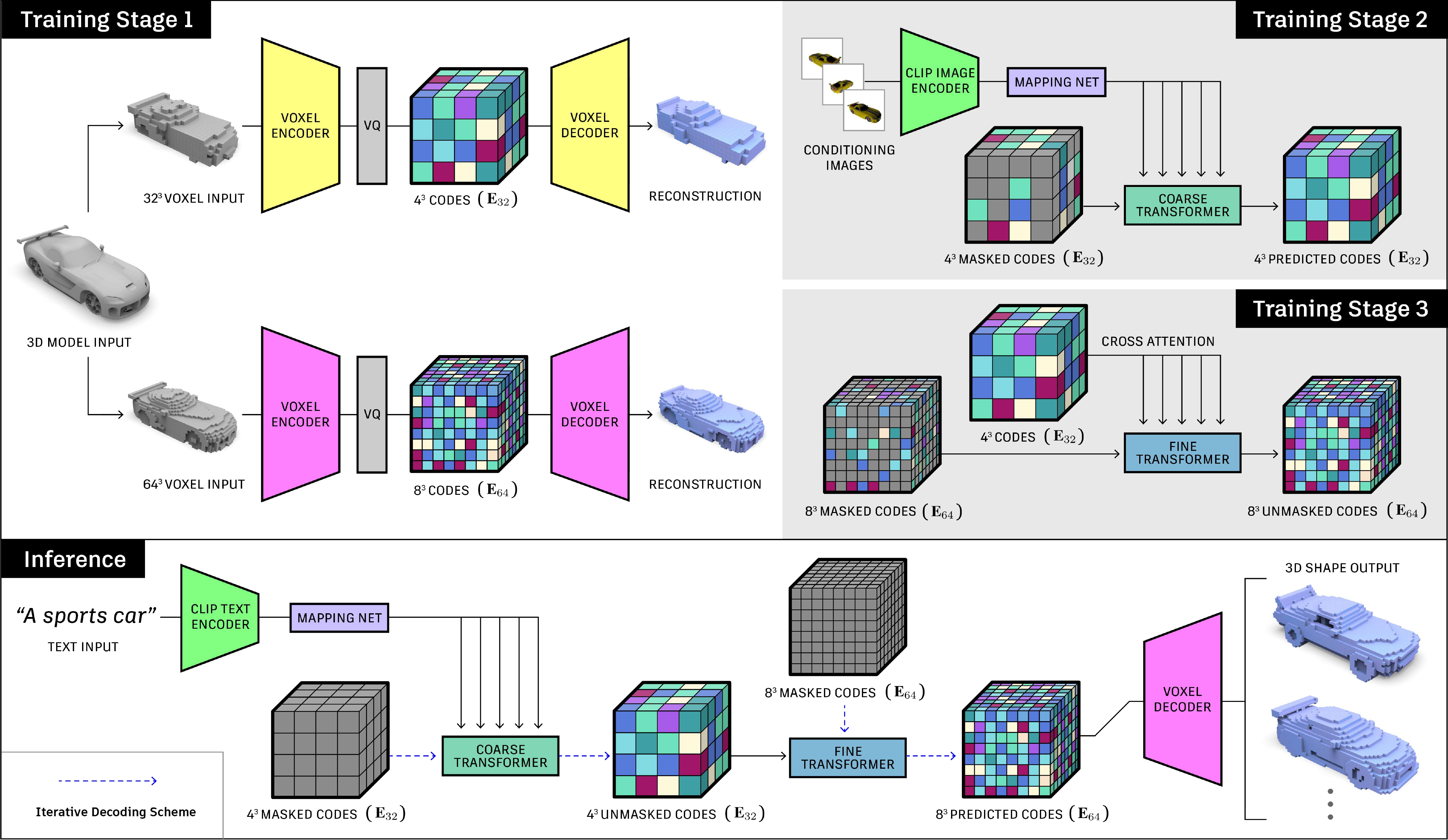}
        \caption{The \nickname~architecture during training (top) and inference (bottom). \nickname{} is trained in three stages. In Stage 1, we train two separate VQ-VAE models for $32^3$ and $64^3$ resolution voxel grids. In Stage 2 we train a coarse transformer conditioned on a CLIP embedding to generate low resolution VQ-VAE latent grids $\textbf{E}_{32}$. In Stage 3, we train a fine transformer to perform super resolution on these latent grids. During inference, a text prompt is passed through the CLIP text encoder and used to condition the coarse transformer to generate a coarse latent grid $\textbf{E}_{32}$. This coarse grid is then used to condition the fine transformer to generate a fine latent grid $\textbf{E}_{64}$. Finally, this fine latent grid is then passed through the Training Stage 1 $64^3$ VQ-VAE decoder to generate the output shape.}

        \label{fig:architecture}
    \end{center}
\end{figure*}
\section{Method}

\nickname aims to generate diverse and high-fidelity 3D shapes of common object categories from text prompts. Figure~\ref{fig:architecture} illustrates the components of our \nickname{} model. 
We train the network using 3 stages. 
In the first stage, we use a vector-quantized representation\cite{oord2017neural}, to efficiently represent 3D shapes in discrete voxel grids. This latent representation captures the diverse semantic meanings of the 3D shapes in common object categories and is compatible with  powerful generative models. To balance computation cost and shape quality, we utilize latent representations at two resolutions. A low resolution latent representation $\textbf{E}_{32}$ encodes geometric information of voxels $\mathbf{V}_{32}$ at resolutions of $32^3$, which is useful for text-conditional generation. A high resolution latent representation $\textbf{E}_{64}$ encodes geometric information of voxel grids $\mathbf{V}_{64}$ at resolutions of $64^3$, which is useful for super-resolution and detail generation. In stage 2,
we train a  generative model conditioned on CLIP features obtained from images $\{ \textbf{I}_r | r \in R\}$ rendered at random views $R$ to solve the lack of  paired (text, shape) data. The conditional generative model consists of a mapping net and a \textit{coarse}  transformer $T_c(.)$ that is trained to unmask a  masked low-resolution grid $\textbf{E}_{32}$. Finally in stage 3,  a \textit{fine} transformer $T_f(.)$ is used as a super-resolution model in latent space to generate higher resolution shapes. It is trained in a similar manner to the \textit{coarse}  transformer but conditioned on a low-resolution grid $\textbf{E}_{32}$ instead of CLIP features. 

During inference, a text prompt is sequentially converted from  CLIP text embedding to a low-resolution latent representation $\textbf{E}_{32}$ to a high-resolution latent representation $\textbf{E}_{64}$ to finally a high-resolution voxel grid $\mathbf{V}_{64}$. A confidence-based iterative decoding scheme \cite{chang2022maskgit} is adopted for unmasking the fully masked initial grids for coarse and fine transformers at inference. To further control shape fidelity and diversity, we propose an annealing strategy for classifier-free guidance at inference time. 

In the following subsections, we will introduce the major components within \nickname. Section \ref{sec:vqvae} describes the vector quantized representations at multiple resolutions; Section \ref{sec:coarsetf} explains the conditional generation model as well as its design choices and training strategies; Section \ref{sec:finetf} introduces the super-resolution model; Section \ref{sec:anneal} explains how the classifier-free guidance is achieved at inference using an annealing strategy.

\subsection{Multi-Resolution Voxel VQ-VAEs}
\label{sec:vqvae}

To learn a low-dimensional latent space that effectively represents the distribution of the training shape set $D$, we use a vector-quantized variational autoencoder (VQ-VAE)~\cite{oord2017neural}. The vector quantized latent representation avoids ``posterior collapse'', and it is also amenable to modeling with transformer-based generative models.

To balance computational cost and shape quality, we train two separate VQ-VAEs for $32^3$ and $64^3$ resolution voxel grids $\mathbf{V}_{32}$ and $\mathbf{V}_{64}$, which are useful for conditional generation and super-resolution respectively.

For both VQ-VAEs, we use ResNet-based\cite{he2016deep} volumetric CNN encoders and decoders. A vector quantization layer maps the down-sampled voxel grids into a discrete latent space by indexing the continuous latent space with an embedding codebook. 
In this way, the voxel grids $\mathbf{V}_{32}$ and $\mathbf{V}_{64}$ are effectively represented by low-dimensional latent representations $\textbf{E}_{32}$ and $\textbf{E}_{64}$, at the resolution of $4^3$ and $8^3$.
The VQ-VAEs are trained with mean squared error loss (MSE), commitment loss\cite{oord2017neural}, and an exponential moving average strategy\cite{razavi2019generating,robust_ema}.

\subsection{CLIP-Conditioned Coarse Transformer}
\label{sec:coarsetf}
The conditional generation process aims to generate diverse shapes that semantically correspond to the text prompts.
Inspired by the recent success of masking approaches~\cite{chang2022maskgit} and diffusion models~\cite{glide, gu2022vector, rombach2022high}, we formulate this generation process as a conditional unmasking task. Given an input masked latent representation $\textbf{E}^{msk}_{32}$ and a conditional vector $\textbf{c}$, the coarse transformer $T_c(.)$ should produce an fully unmasked latent representation $\textbf{E}_{32}$ that not only aligns with the semantic meaning captured by the conditional vector, but also allows for the sampling of diverse shapes at inference. 


The conditional vector $\textbf{c}$ provides the semantic guidance for the shape generation process. Since we assume no text data at the training time, we leverage the text-image joint-embedding of CLIP. The generation is guided by the CLIP image embedding from the rendered images $\{ \textbf{I}_r\}$ during training and guided by the CLIP text embedding from the text prompt during inference.
We add Gaussian noise to the CLIP image embeddings~\cite{zhou2021lafite} to alleviate the modality gap between the CLIP text and image embeddings~\cite{liang2022mind}:
\begin{equation}
  \mathbf{\hat{c}} =  f_{I}(\textbf{I}_r) +  \frac{\gamma \cdot \epsilon \cdot \lVert f_{I}(\textbf{I}_r) \rVert_2}{\lVert  \epsilon\rVert_2} ,
\label{eq:condition_vector}
\end{equation}
where $f_{I}(\cdot)$ is the CLIP image encoder, $\epsilon \sim \mathcal{N}(0, 1) $, and $\gamma$ controls the level of perturbation.
The conditional vector $\mathbf{\hat{c}}$ is then normalized and passed through a mapping network that consists of multi-layer perceptrons (MLP).
The mapped vector $\mathbf{\tilde{c}}$ conditions on the coarse transformer $T_c(.)$ by predicting the affine transform parameters of each transformer block's layernorm. 
In order to control the diversity-quality balance of the generated shapes, we replace the conditional vector with the null embedding by a $\rho\%$ drop out rate for classifier-free guidance~\cite{ho2022classifier}. These are important design choices that affects the diversity and quality of the generated shapes at inference time. 

To train  the coarse transformer $T_c(.)$, we randomly replace the input indices with a special ``mask'' tokens by uniformly selecting a masking ratio between 0-100\% \cite{chang2022maskgit}. Given the masked latent grid $\mathbf{E}^{msk}_{32}$ and the conditional vector $\mathbf{c}$, the training objective is to maximize the following log-likelihood: $\sum_n^N \log p(\mathbf{E}_{32,n}|\mathbf{E}^{msk}_{32,n}, \mathbf{c})/N$, where $N$ is the training data size and $\mathbf{E}^{msk}_{32}$ is the randomly masked latent grids. 
As the network sees only the ground truth tokens but not its own predicted samples during training, there exists an autoregressive drift resulted by the accumulative error at inference-time sampling. To alleviate this issue, we take inspiration from the NLP literature~\cite{savinov2021step} and use a similar step-unrolled training (SUT) that predicts the entire unmasked grids at each time step. The final objective is to maximize the following formulation:
\begin{equation}
 \sum_n^N \Big(\log p(\mathbf{E}_{32,n}|\mathbf{E}^{msk}_{32,n}, \mathbf{c}) + \log p(\mathbf{E}_{32,n}|\tilde{\mathbf{E}}_{32,n}, \mathbf{c}) \Big) \Big/ N,
\end{equation}
where $\tilde{\mathbf{E}}_{32}$ is the coarse transformer $T_c(.)$'s own prediction. 

\subsection{Super-Resolution with Fine Transformer}
\label{sec:finetf}
Super-resolution in the discrete latent space improves quality and upsamples the  resolution of generated shapes. This process is done with a fine transformer $T_f(.)$, which takes a low-resolution latent representation $\mathbf{E}_{32}$ and a masked high-resolution latent grid as input. The fine transformer is conditioned on the coarse grid $\mathbf{E}_{32}$ via cross attention. It unmasks the high-resolution representation $\mathbf{E}_{64}$.
To ensure that training data resembles the inference data, we directly use the predictions $\tilde{\mathbf{E}}_{32}$ of the coarse transformer $T_c(.)$ to train the fine transformer $T_f(.)$, instead of using the ground truth low-resolution latent representation $\mathbf{E}_{32}$. 

\subsection{Annealed Classifier-Free Guidance}
\label{sec:anneal}
At inference time, we first convert an input text prompt into a CLIP text embedding using the CLIP text encoder $f_T(\cdot)$. The text embedding is then fed into the mapping network and used as the conditional vector $\mathbf{\tilde{c}}$ of the coarse transformer $T_c(.)$. 
We use the iterative decoding scheme \cite{chang2022maskgit} to slowly unmask the grid over a sequence of $T$ steps. The initial input to the coarse transformer $T_c(.)$ is a completely masked latent grid. At each time step $t$, we condition the coarse transformer $T_c(.)$ with $\mathbf{\tilde{c}}$ and take the output latent grid $\mathbf{E}^{msk}_{32, t-1}$ from the prior time step as the input. 
The coarse transformer $T_c(.)$ will predict an unmasked latent grid $\tilde{\mathbf{E}}_{32,t}$. We mask all the tokens except for the most confident token predictions and the tokens unmasked from previous steps. This newly masked latent grid $\mathbf{E}^{msk}_{32, t}$ is the input to the next time step $t+1$. 
We repeat this process until the process unmasks all the tokens, which is ensured by a cosine masking schedule \cite{chang2022maskgit}.

During this iterative decoding scheme, we apply a new variant of classifier-free guidance \cite{ho2022classifier}. Classifier-free guidance extrapolates an unconditional sample in the direction of a conditional sample. The extrapolation is controlled by an adaptive \emph{guidance scale} parameter. Varying this parameter results in a trade-off between accuracy of text-to-shape generation and sample diversity. 
Instead of using a constant guidance scale for all time steps~\cite{ho2022classifier}, we propose to vary the guidance scale over time according to an annealing schedule. 
The intuition is that during the initial steps of sampling when the input to the coarse transformer is mostly masked indices, a larger guidance scale is important to keep the network ``on task''; later on in the sampling process, a lower guidance scale can help produce more sample diversity. Overall, the guidance is formulated as:

\begin{equation}
\begin{aligned}
  \hat{P}_{t}(c) = P_{t}(0)  + a(t) \Big( P_{t}(c) - P_{t}(0)\Big),
  \end{aligned}
\end{equation}
where $a(t)$ is the guidance scale annealing schedule, $P_{t}(c)$ denotes the conditional distribution $p(\mathbf{E}_{32,t}|\mathbf{E}^{msk}_{32, t-1}, \mathbf{c})$, and $P_{t}(0)$ denotes $p(\mathbf{E}_{32,t}|\mathbf{E}^{msk}_{32, t-1}, \mathbf{0})$.
Note that $a(t)$ is continuous and usually monotonically decreasing. By setting $a(t) = k$ for some constant $k$, this equation is equal to the classifier-free guidance scheme with a constant guidance scale. 
In this paper, we experimentally evaluate other annealing schedules~\cite{chang2022maskgit}, including linear, cosine, and square root functions.  

Finally, the unmasked coarse latent grid $\tilde{\mathbf{E}}_{32}$ is fed to the fine transformer $T_f$, which uses the iterative decoding scheme to unmask an initially 
 masked fine latent grid $\textbf{E}^{msk}_{64}$.
The unmasked fine grid is then passed to the $64^3$ VQ-VAE decoder to obtain the final high-resolution voxel shape.

\section{Experiments}
\label{sec:experiments}
In this section, we report experimental results to evaluate the generation fidelity, diversity, and class accuracy of \nickname.  We provide full details of the experiment setup in the supplementary. We run the experiments 3 times for each of the aforementioned measures and report the mean in each case, except for the final comparisons with baselines (Table \ref{tab:main_com}), where we use the best seed. Additional results can also be found in the supplementary.

\noindent \textbf{Dataset. }
 We conduct our experiments on two subsets of the ShapeNet(v2) dataset \cite{chang2015shapenet}. The first subset, \textit{ShapeNet13}, contains 13 categories from ShapeNet as used in  \cite{choy20163d,OccupancyNetworks2019}. We use the same train/test split as specified in \cite{OccupancyNetworks2019}. Our second subset is  \textit{ShapeNet55} which contains all 55 ShapeNet categories. For ShapeNet55, we render images as described in \cite{choy20163d}; the training dataset contains 51,784 datapoints, whereas the test set contains 6,101 datapoints.

\noindent \textbf{Evaluation Metrics. }

 To assess the generative capabilities of \nickname{}, we use Fréchet Inception Distance (\textbf{FID}) (\cite{heusel2017gans}) to evaluate shape diversity and quality, and use Classifier Accuracy (\textbf{Acc}) to evaluate the text-shape correspondence. These metrics follow~\cite{sanghi2022clip} where they first generated single mean shapes for 234 predetermined text queries and then passed all shapes through a classifier to measure \textbf{Acc}. The latent space of this classifier is also used for \textbf{FID}.

\newcolumntype{P}[1]{>{\centering\arraybackslash}p{#1}}
\begin{figure*}[t!]
\centering
\setlength{\tabcolsep}{1pt}
\scriptsize
\begin{tabular}{P{3.48cm}P{1.74cm}P{1.74cm}P{3.48cm}P{1.74cm}P{1.74cm}P{1.74cm}P{1.74cm}}

\multicolumn{3}{c}{\textbf{\nickname}} & \multicolumn{3}{c}{\textbf{CLIP-Forge}} & {\textbf{DreamField}}& {\textbf{CLIP-Mesh}}\\
\includegraphics[width=1.00\linewidth]{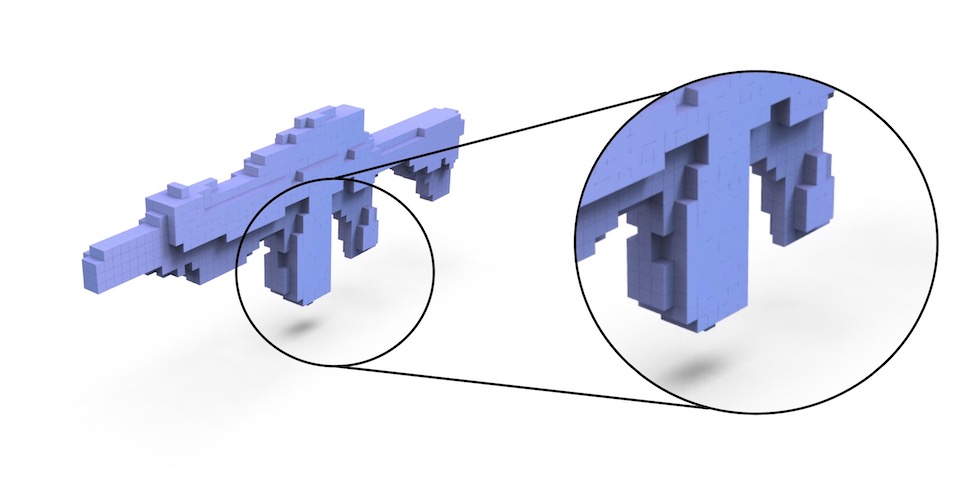} &
\includegraphics[trim={3cm 3cm 3cm 3cm},clip,width=1.00\linewidth]{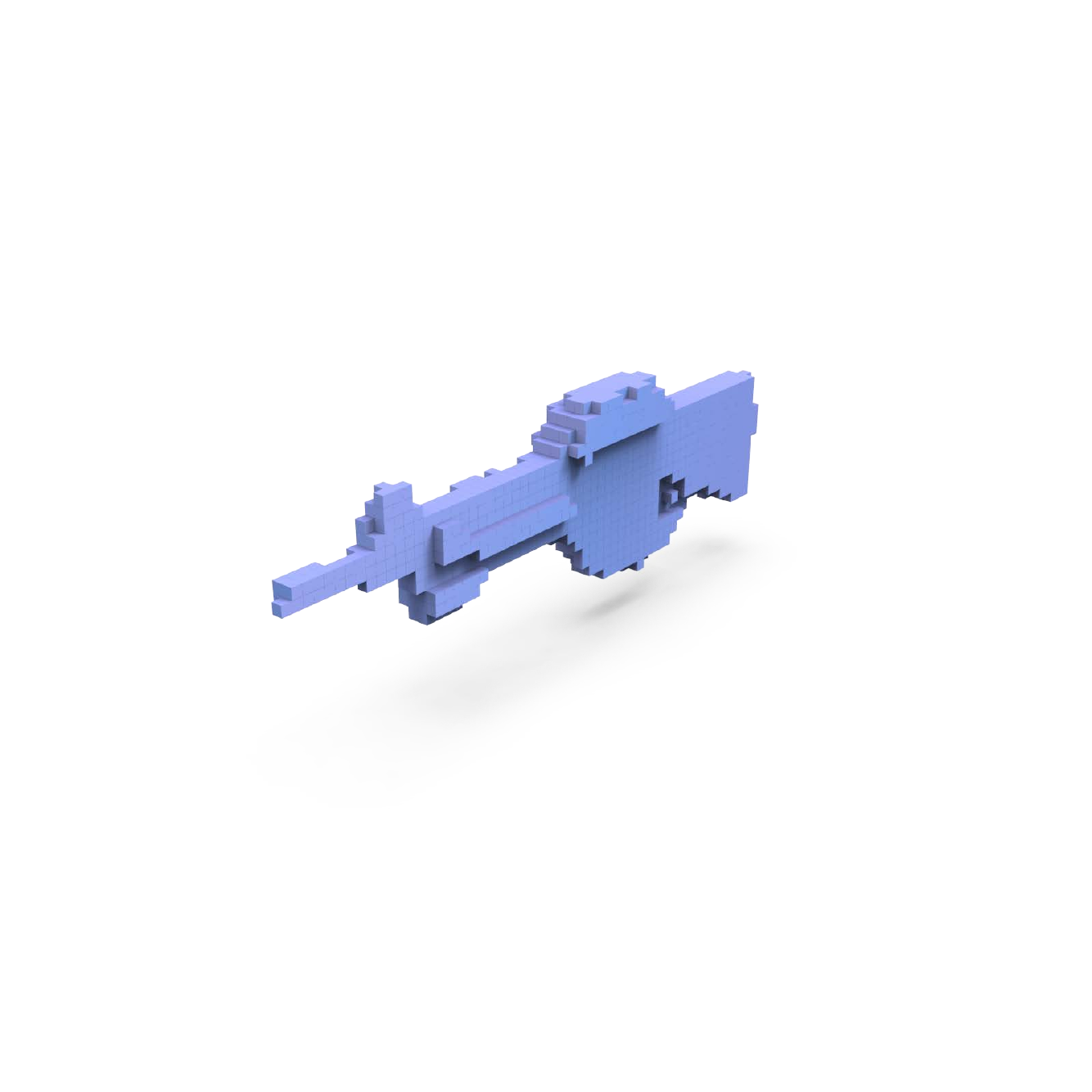} &
\includegraphics[trim={3cm 3cm 3cm 3cm},clip,width=1.00\linewidth]{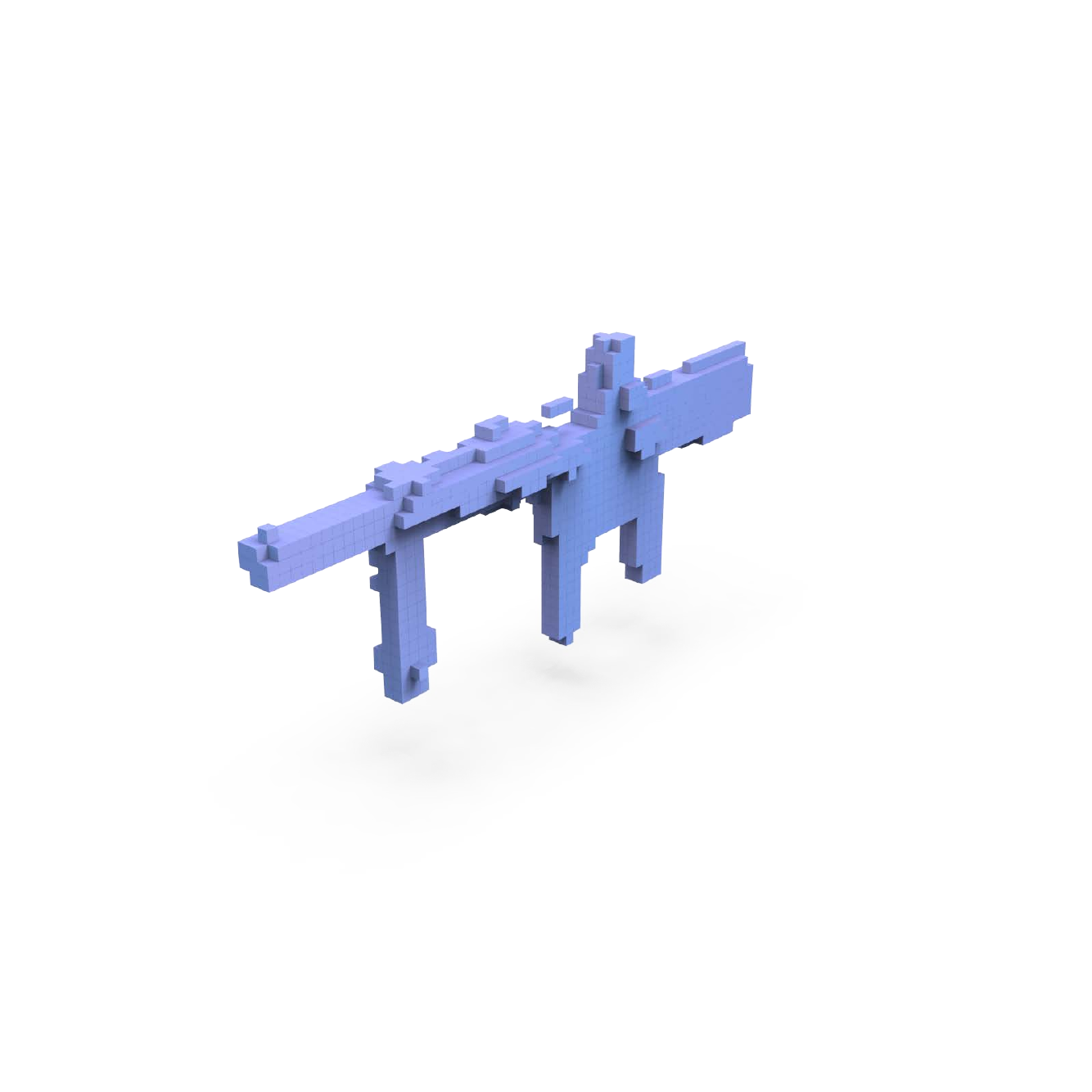} &
\includegraphics[width=1.00\linewidth]{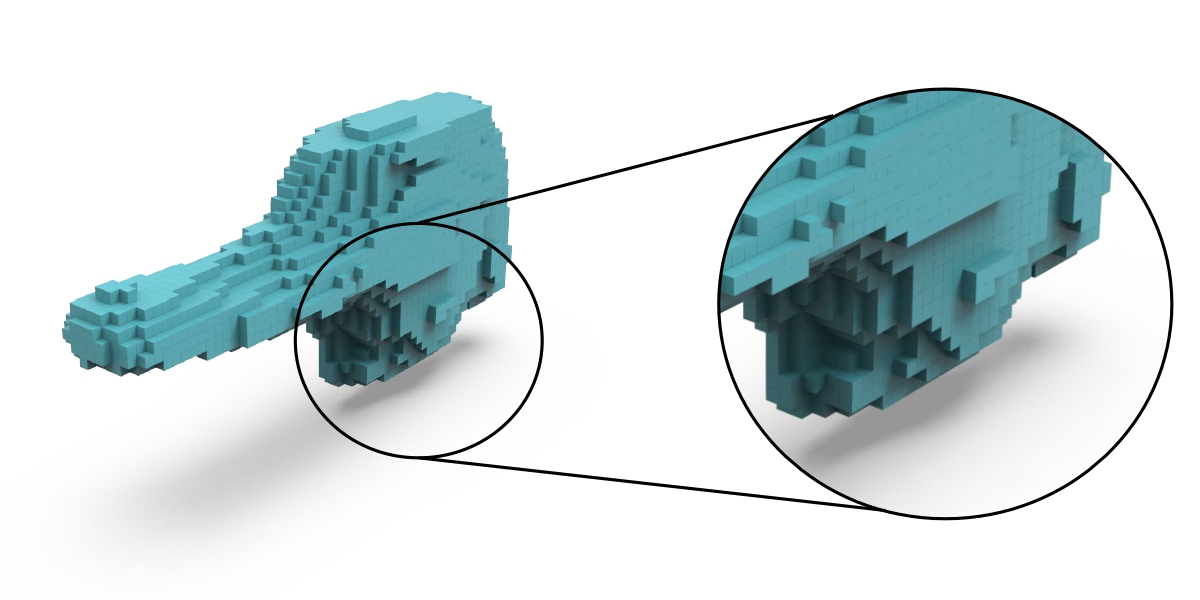} &
\includegraphics[width=1.00\linewidth]{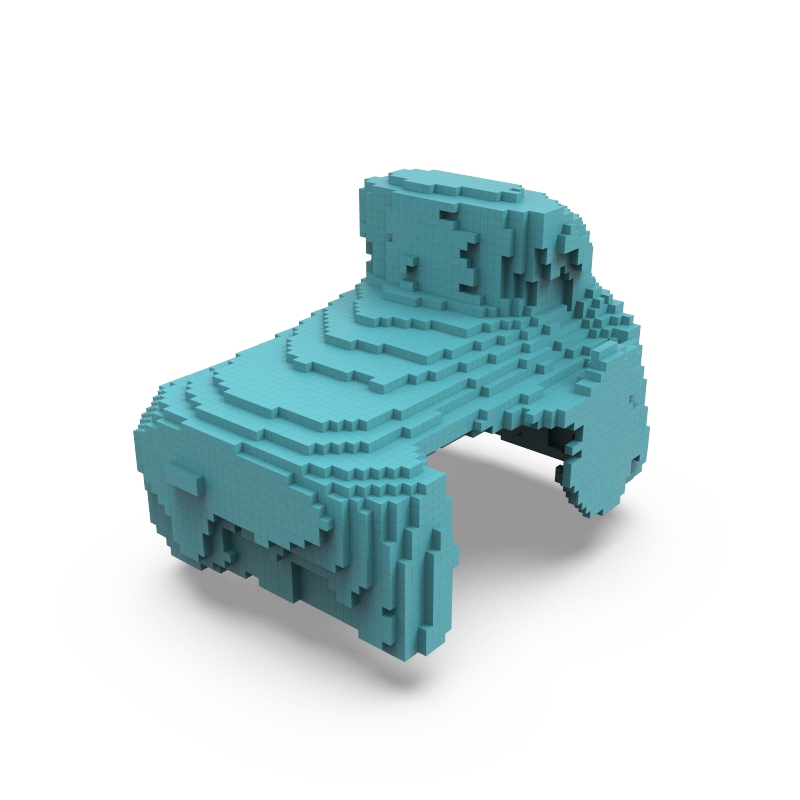} &
\includegraphics[width=1.00\linewidth]{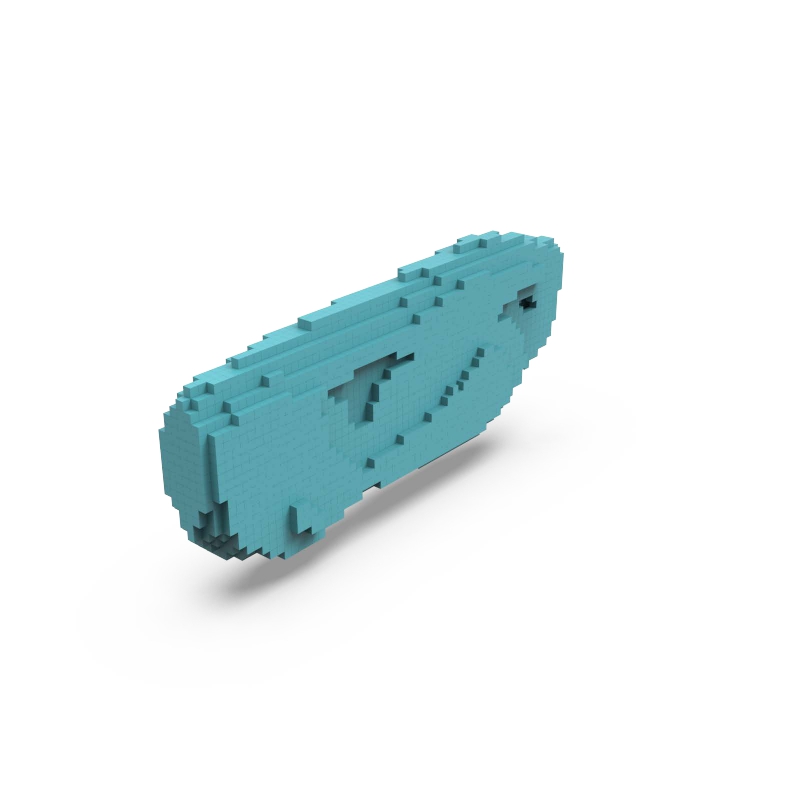} &
\includegraphics[trim={2.5cm 2.5cm 2.5cm 2.5cm},clip,width=1.00\linewidth]{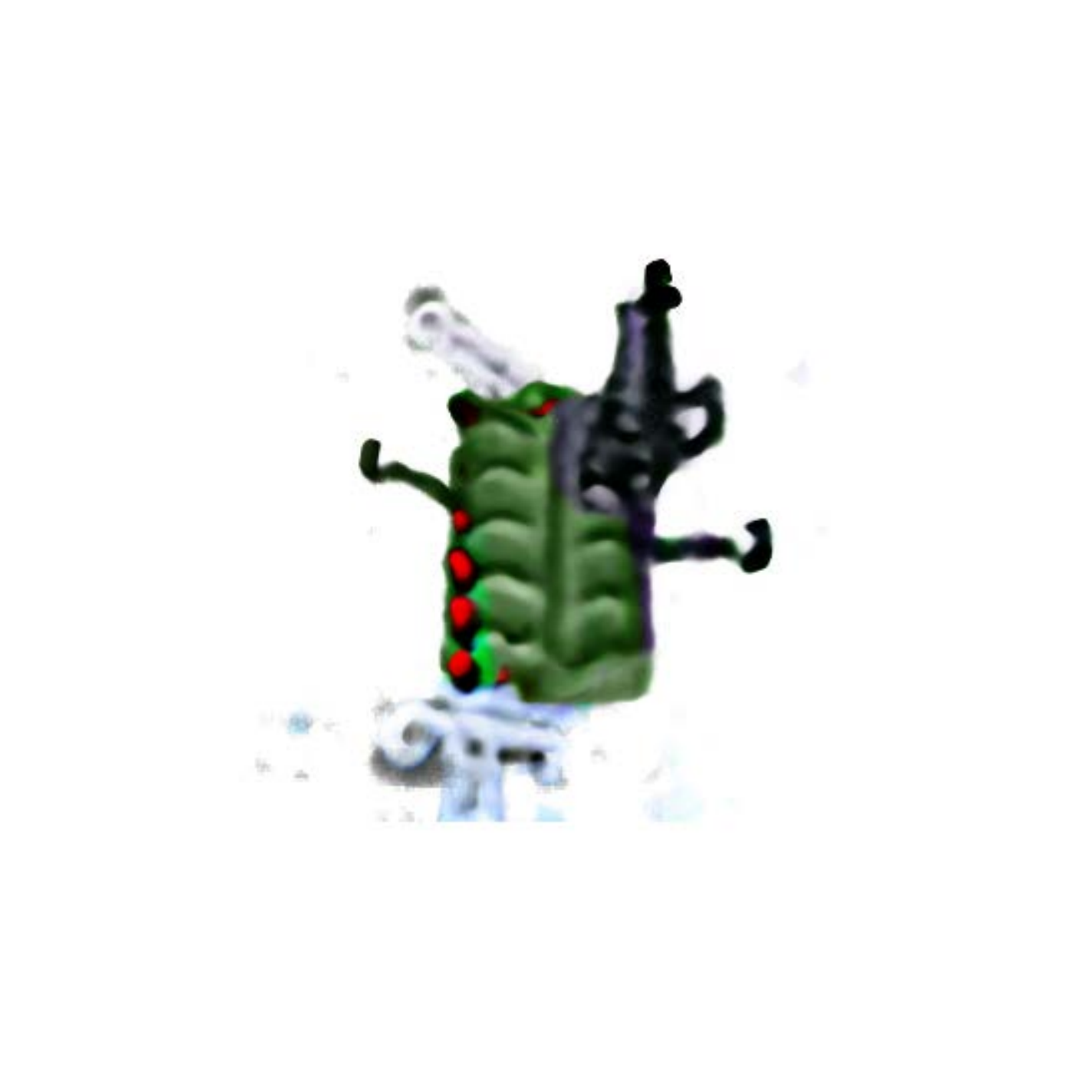} &
\includegraphics[trim={3cm 3cm 3cm 3cm },clip,width=1.00\linewidth]{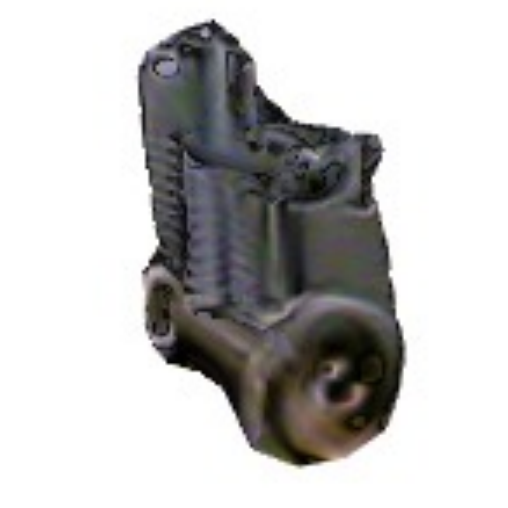}\\
\multicolumn{8}{c}{\small{\texttt{"a machine gun"}}}
\\

\includegraphics[width=1.00\linewidth]{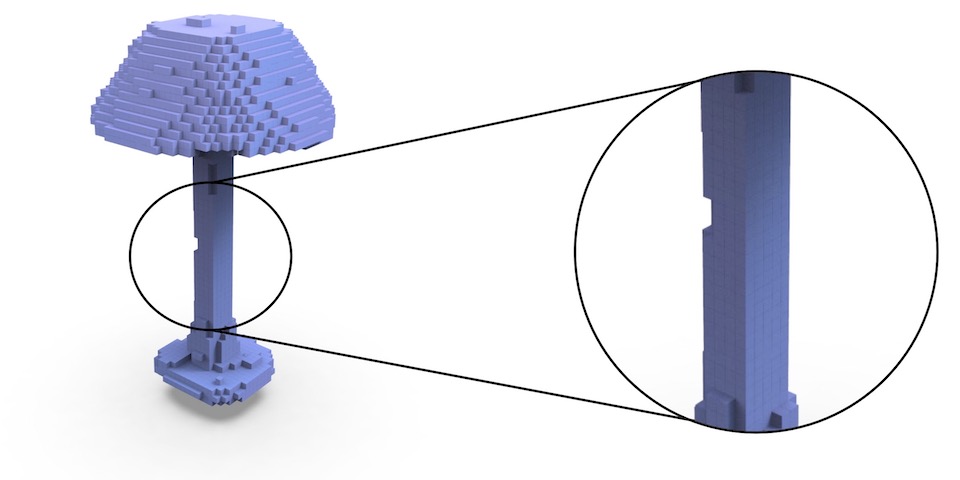} &
\includegraphics[trim={1.5cm 1.5cm 1.5cm 1.5cm},clip,width=1.00\linewidth]{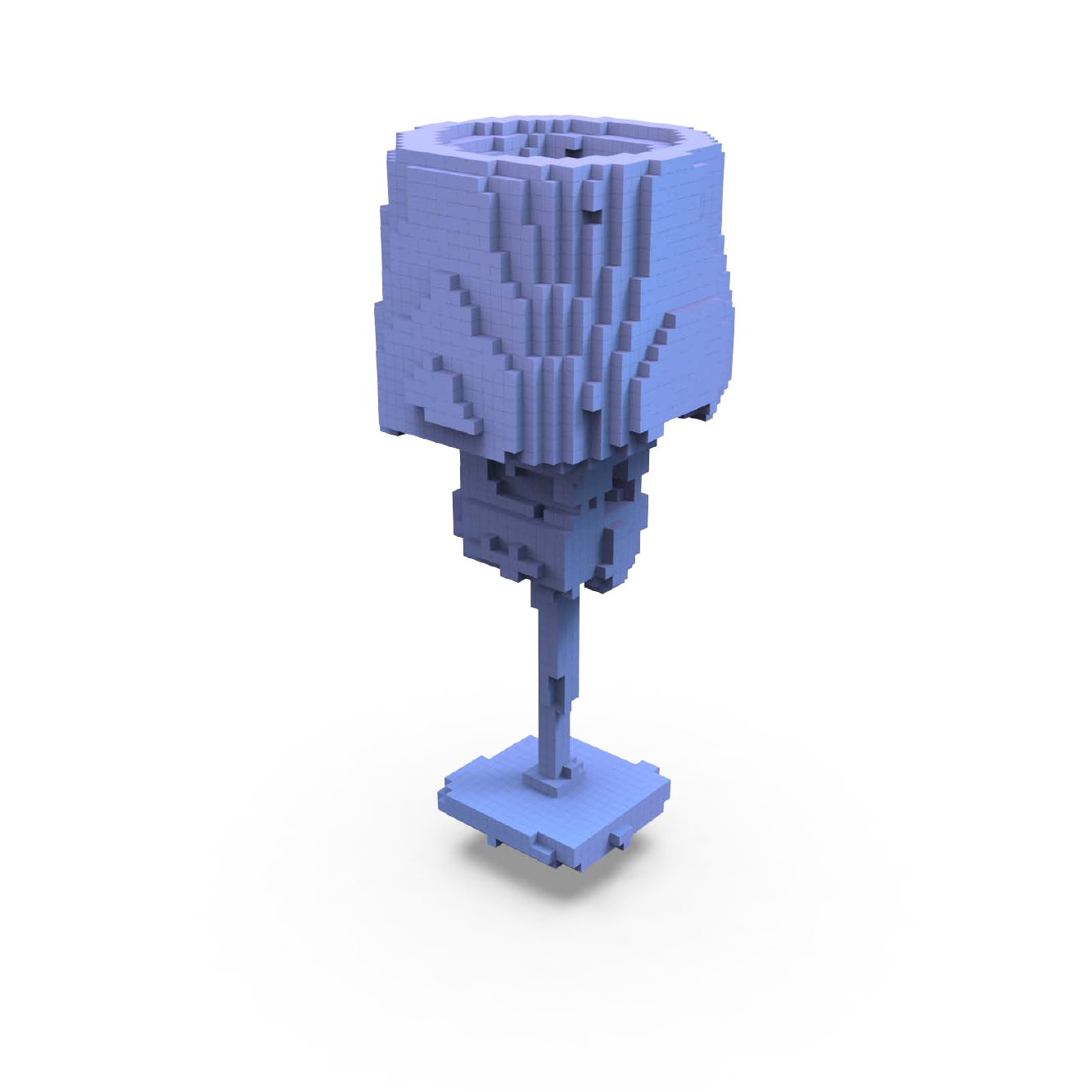} &
\includegraphics[trim={1.5cm 1.5cm 1.5cm 1.5cm},clip,width=1.00\linewidth]{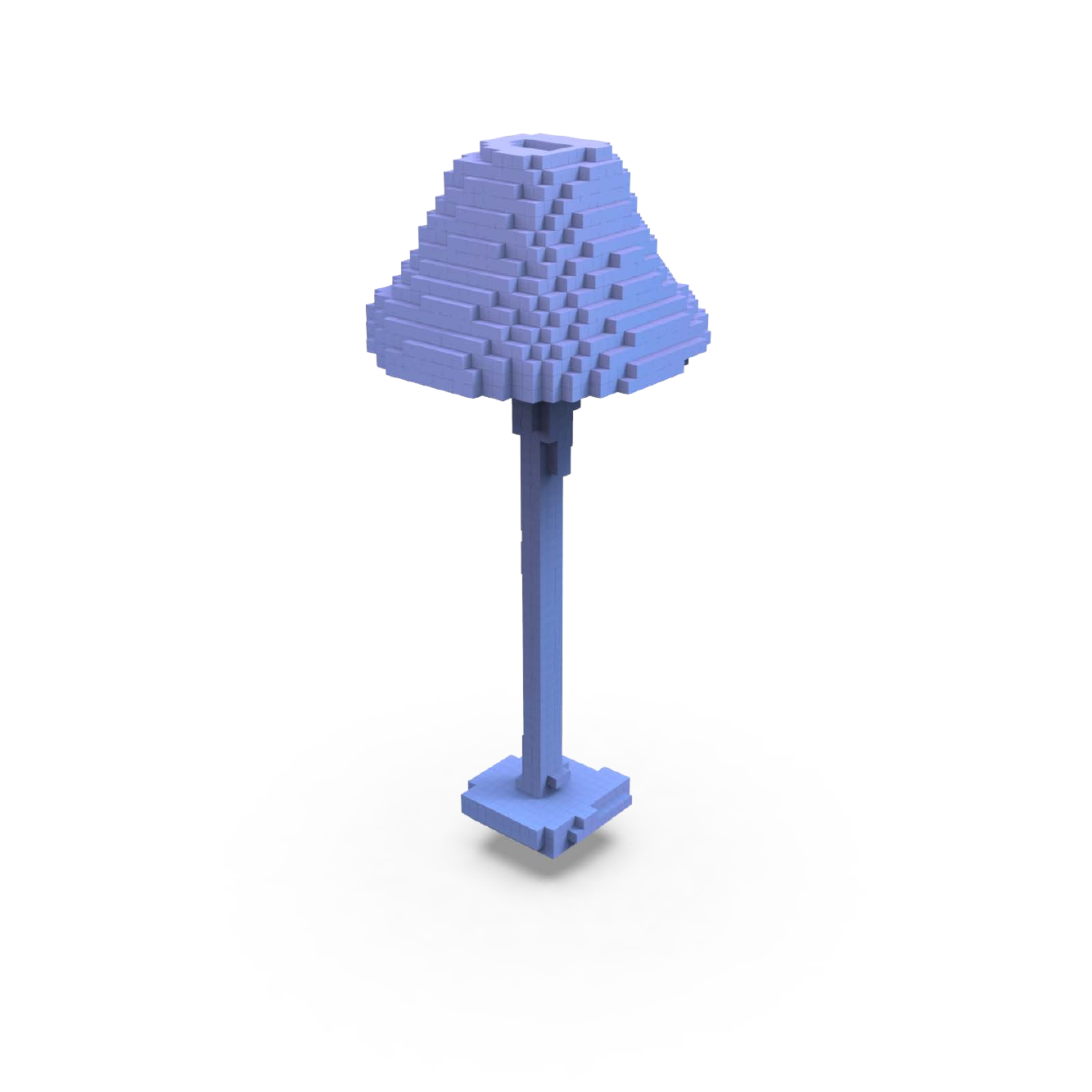} &
\includegraphics[width=1.00\linewidth]{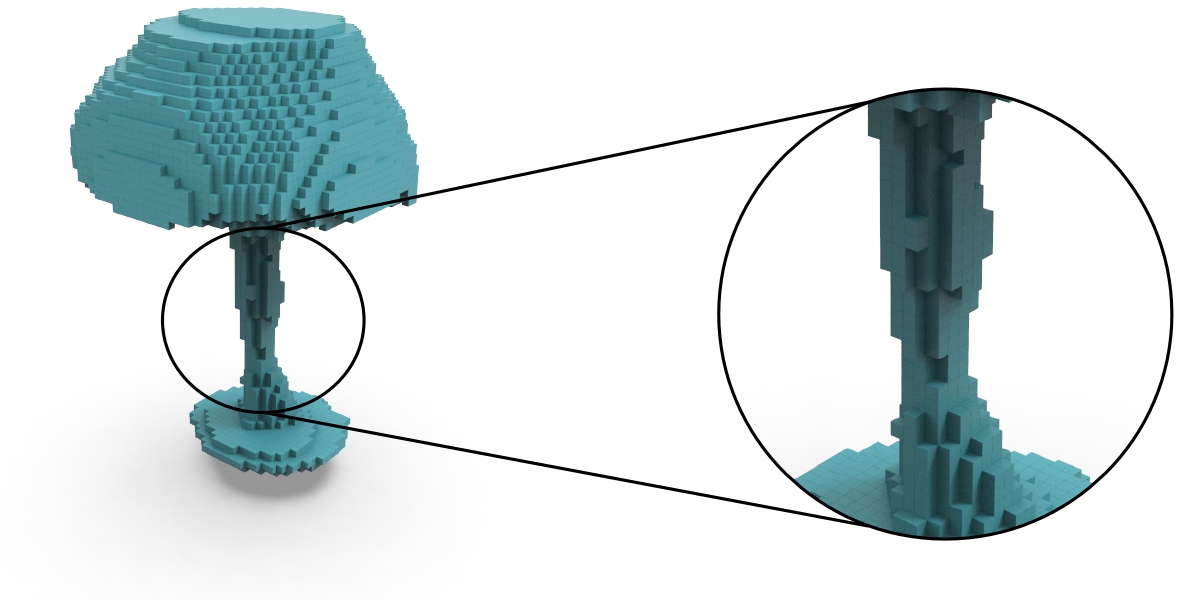} &
\includegraphics[width=1.00\linewidth]{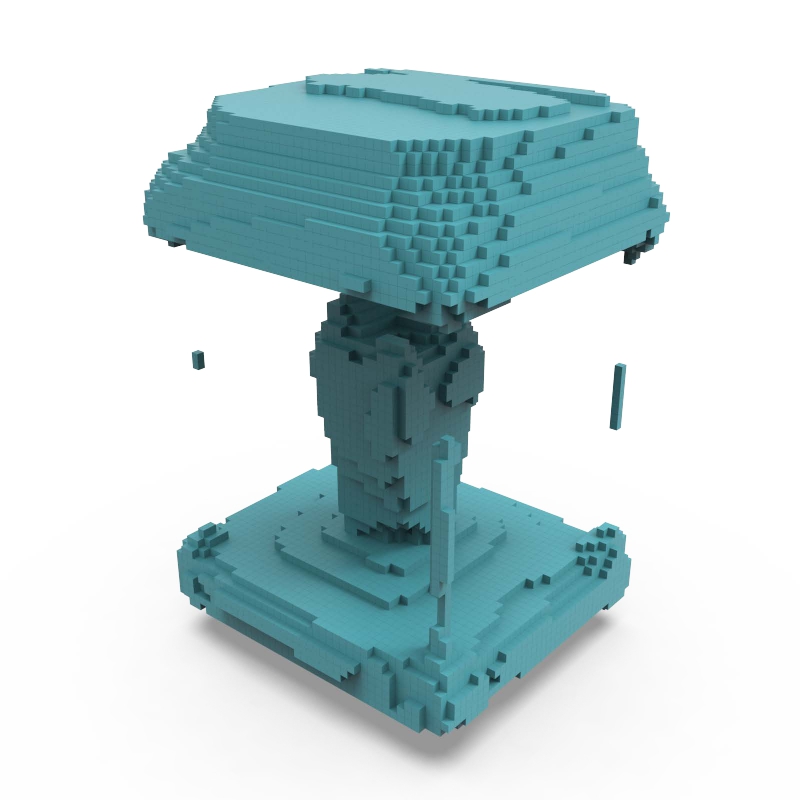} &
\includegraphics[width=1.00\linewidth]{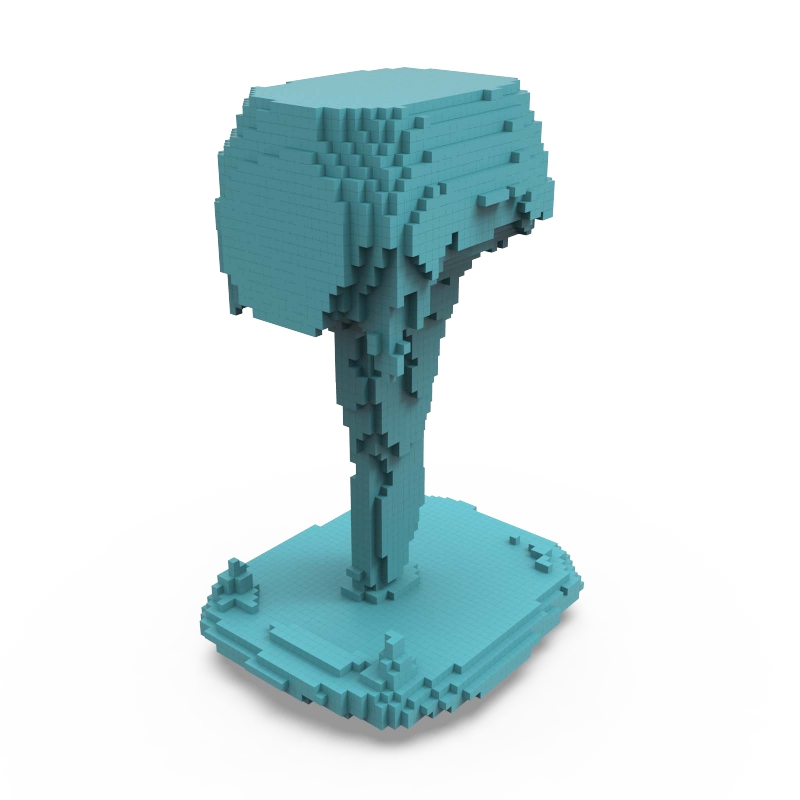} &
\includegraphics[trim={2.5cm 2.5cm 2.5cm 2.5cm},clip,width=1.00\linewidth]{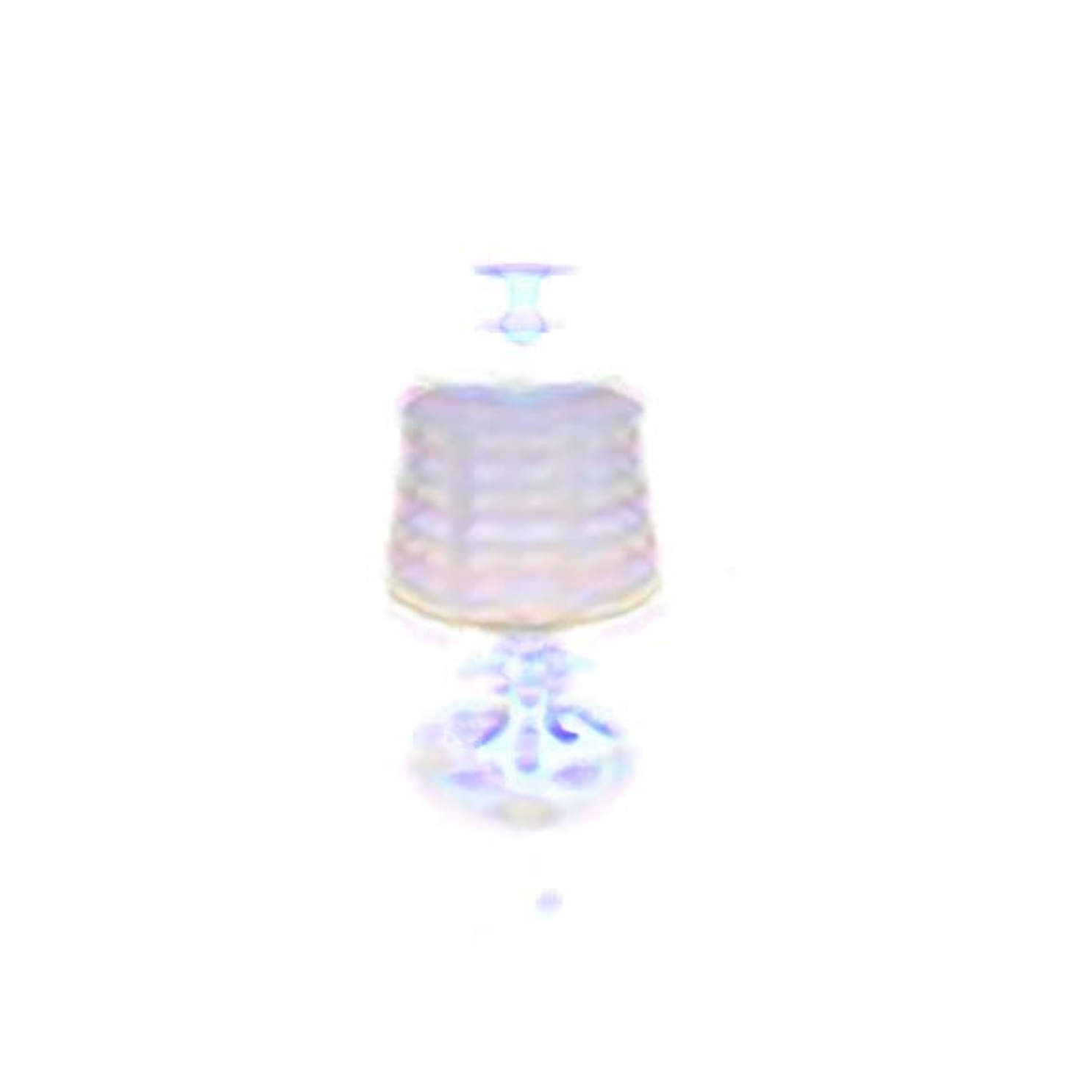} &
\includegraphics[trim={3cm 3cm 3cm 3cm },clip,width=1.00\linewidth]{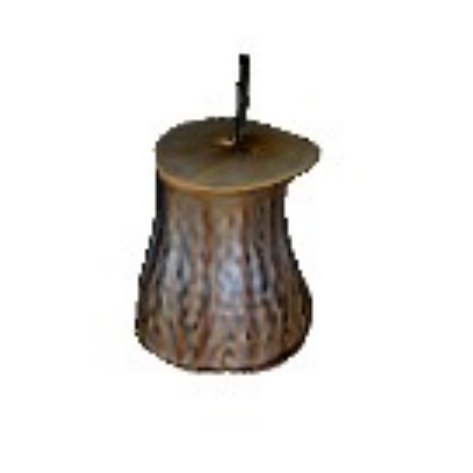}\\
\multicolumn{8}{c}{\small{\texttt{"a table lamp"}}}
\\

\includegraphics[width=1.00\linewidth]{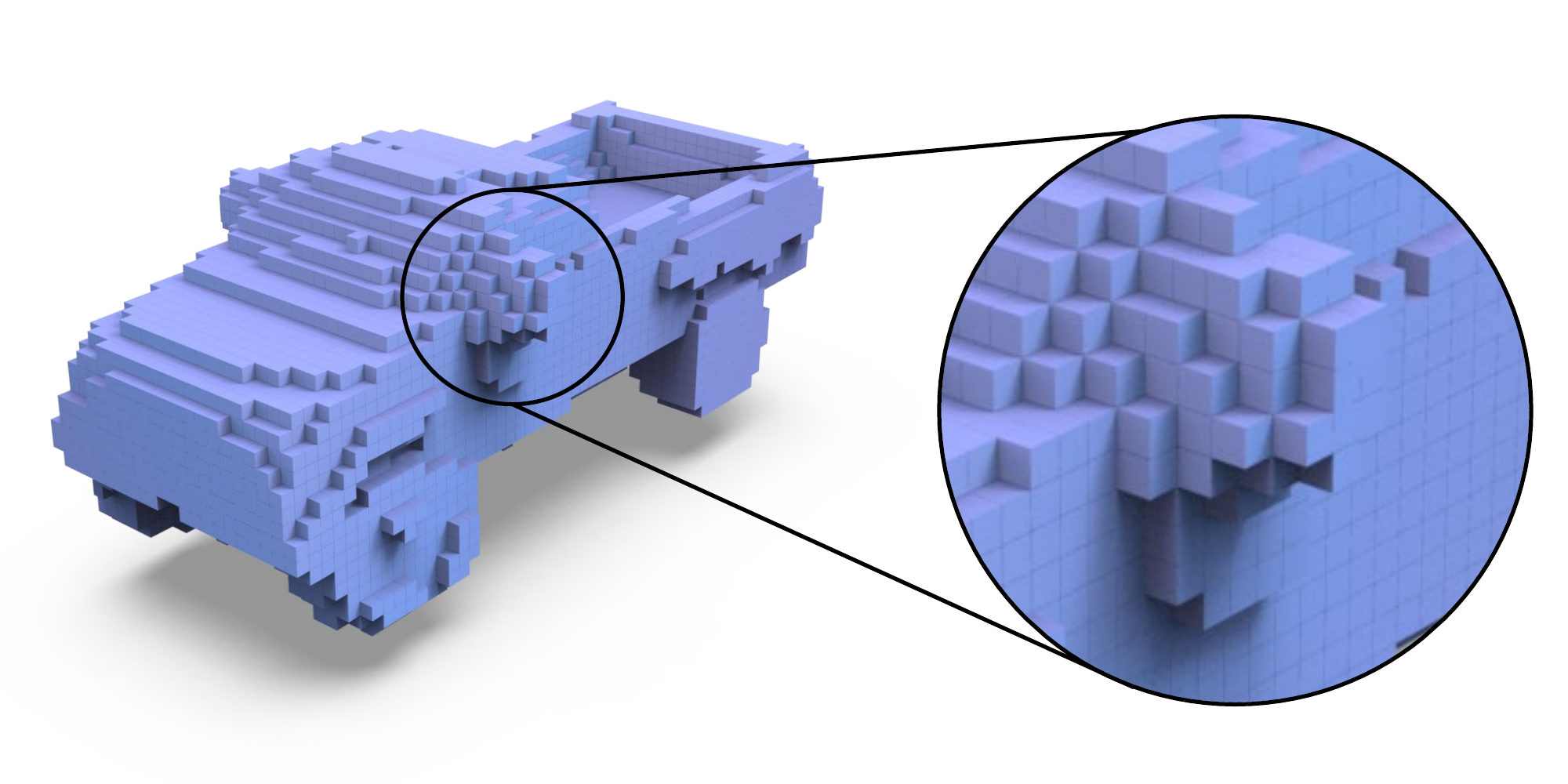} &
\includegraphics[trim={2.2cm 2.2cm 2.2cm 2.2cm},clip,width=1.00\linewidth]{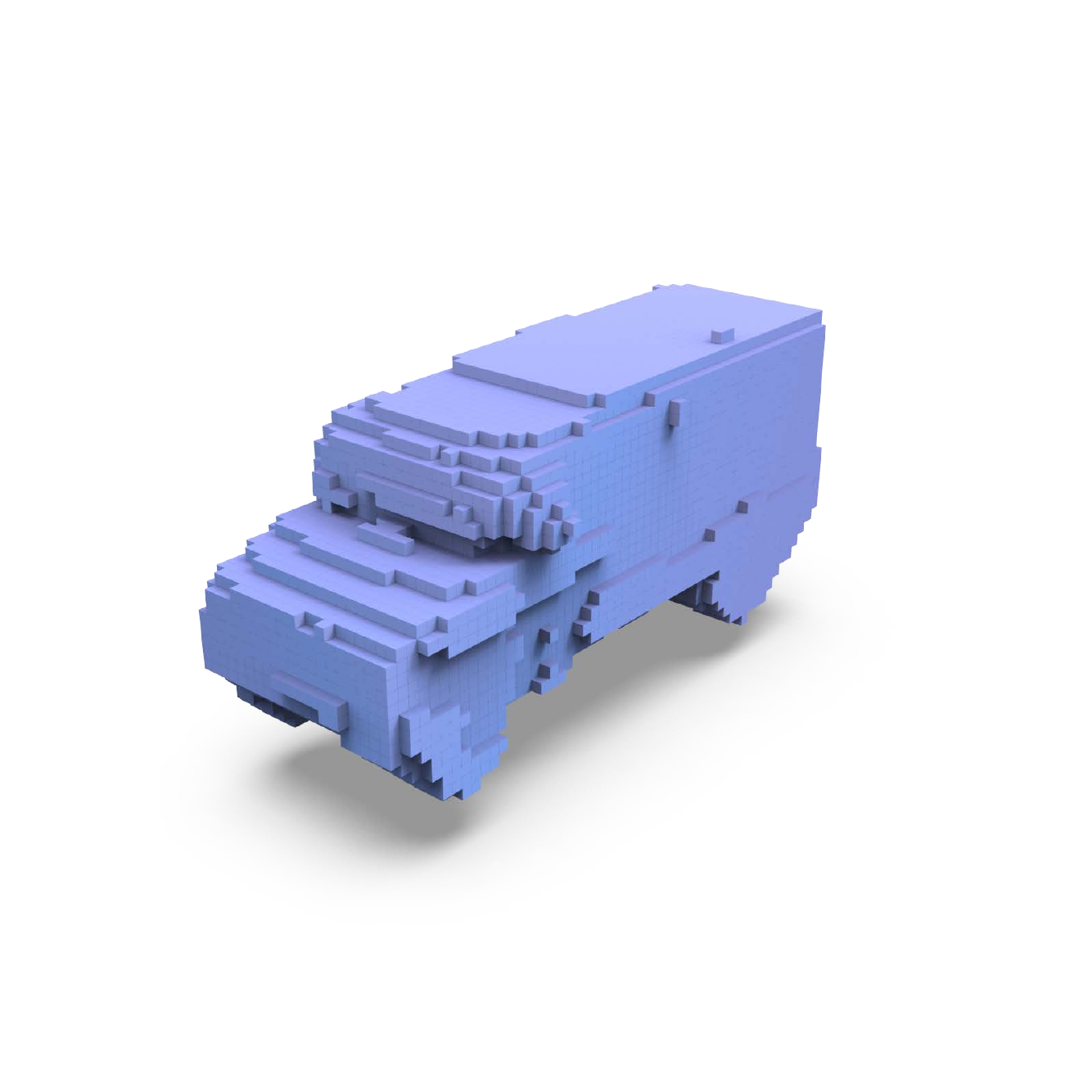} &
\includegraphics[trim={2.2cm 2.2cm 2.2cm 2.2cm},clip,width=1.00\linewidth]{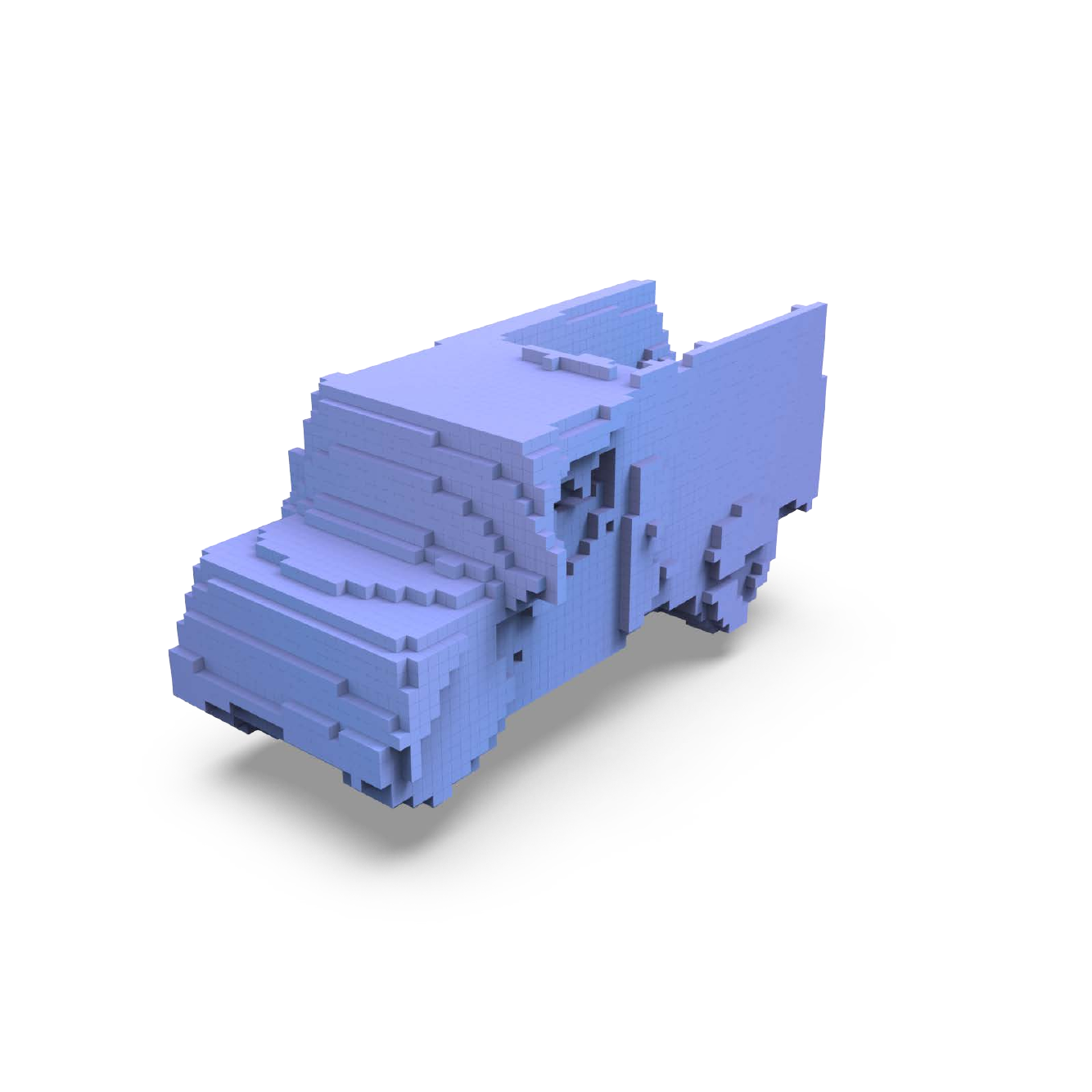} &
\includegraphics[width=1.00\linewidth]{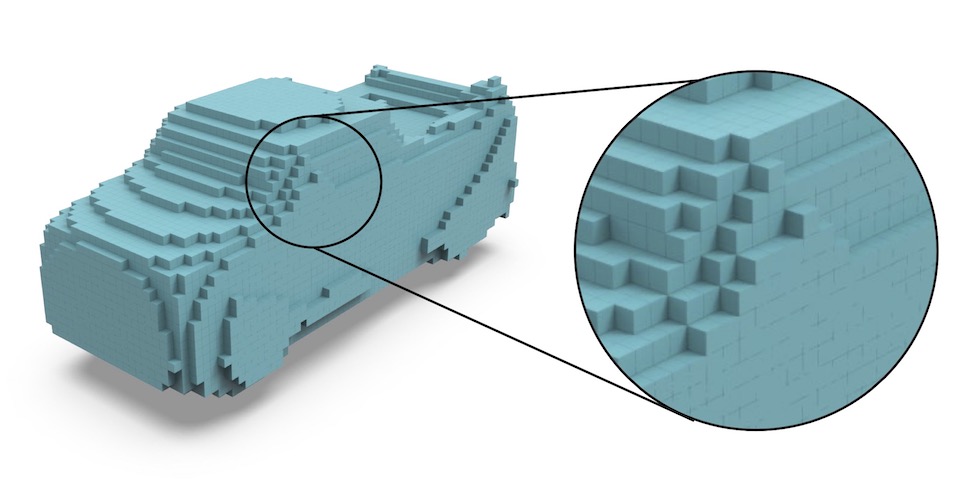} &
\includegraphics[trim={1cm 1cm 1cm 1cm},clip,width=1.00\linewidth]{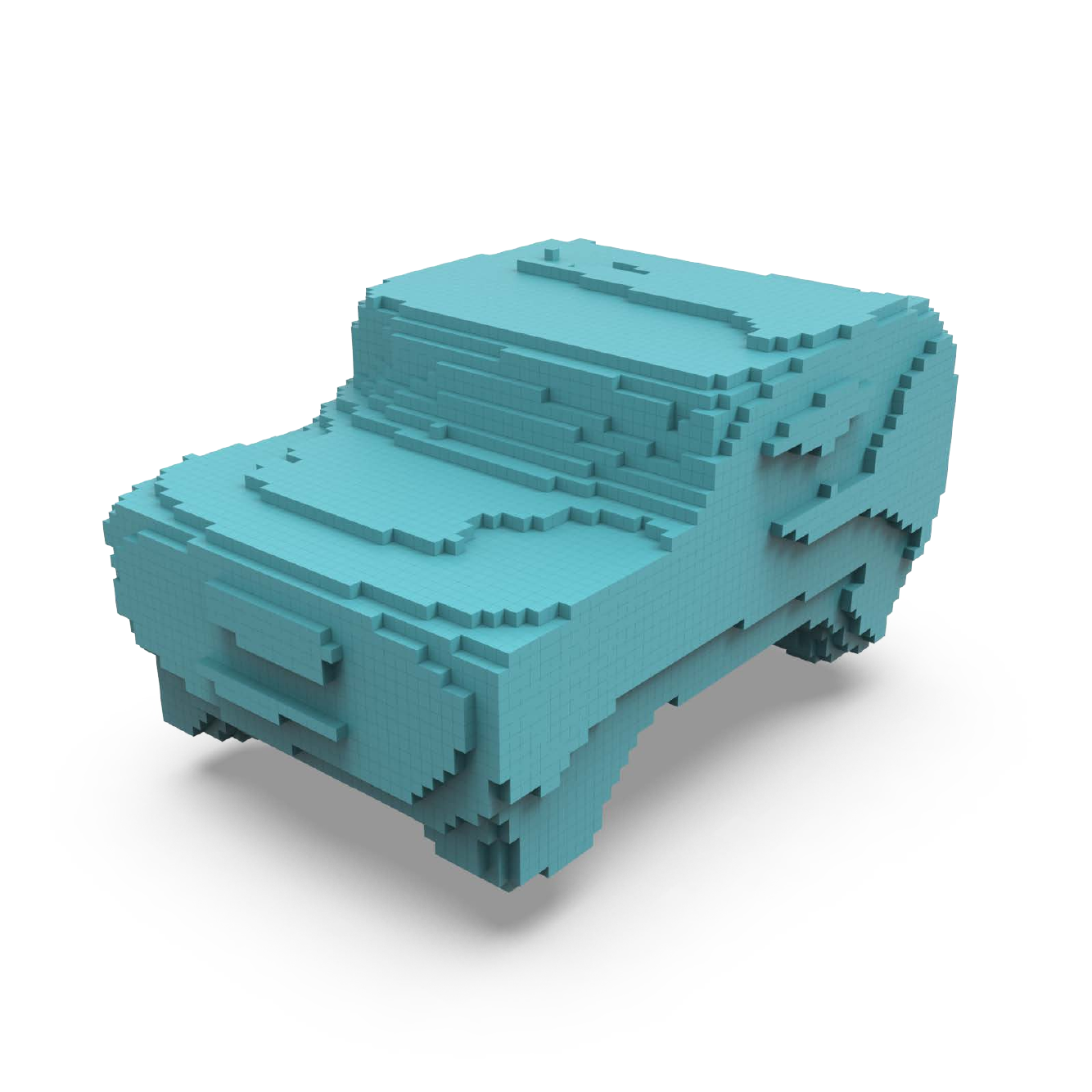} &
\includegraphics[trim={1cm 1cm 1cm 1cm},clip,width=1.00\linewidth]{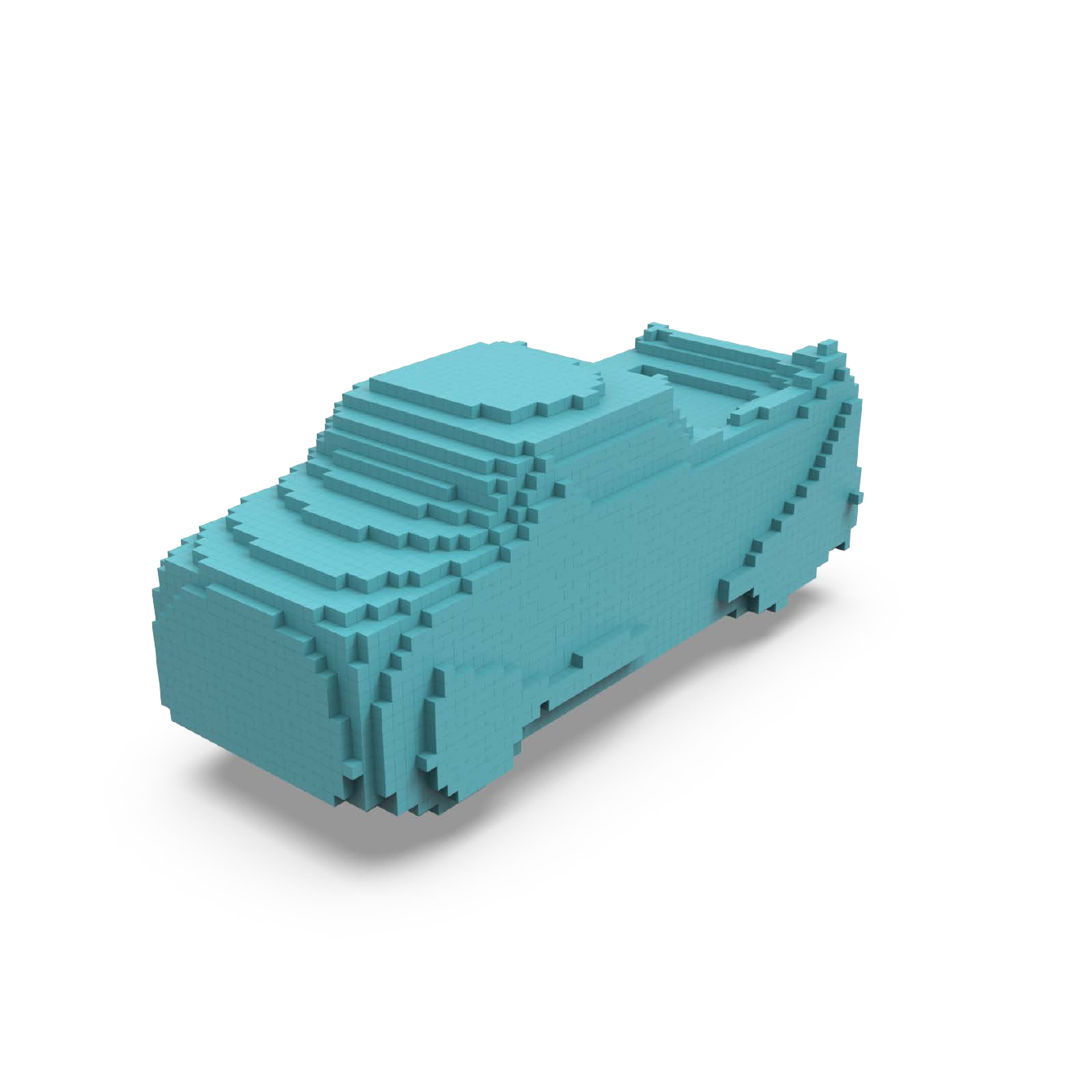} &
\includegraphics[trim={2.5cm 2.5cm 2.5cm 2.5cm},clip,width=1.00\linewidth]{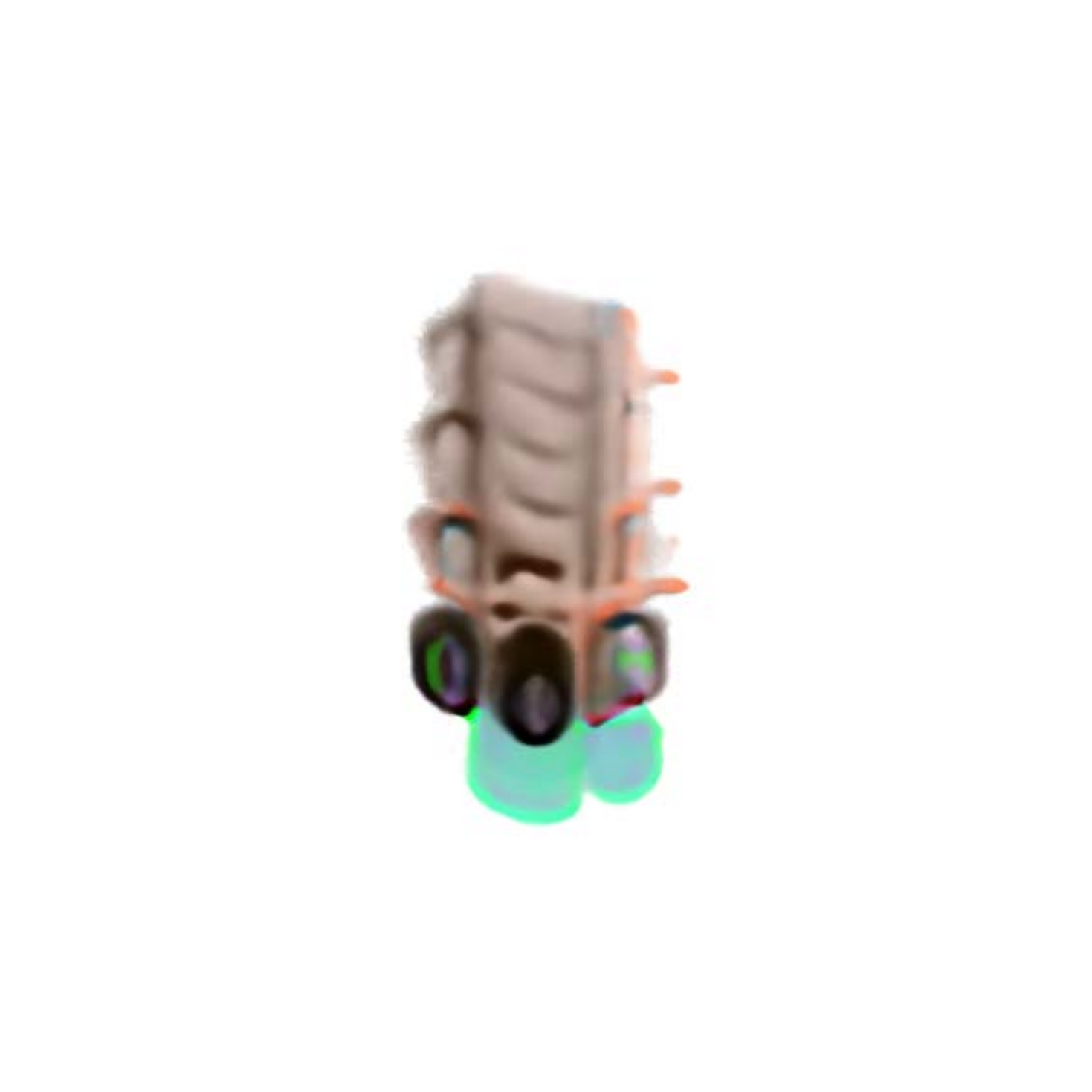} &
\includegraphics[trim={3cm 3cm 3cm 3cm },clip,width=1.00\linewidth]{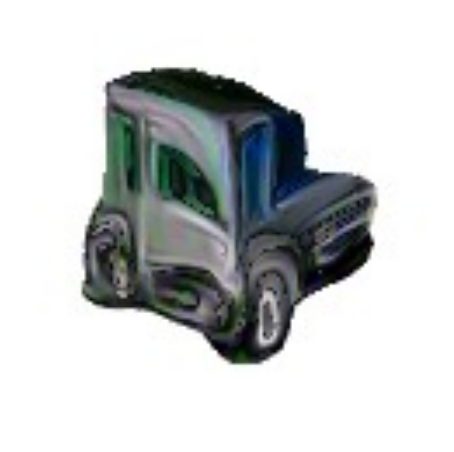}\\
\multicolumn{8}{c}{\small{\texttt{"a truck"}}}
\\

\includegraphics[width=1.00\linewidth]{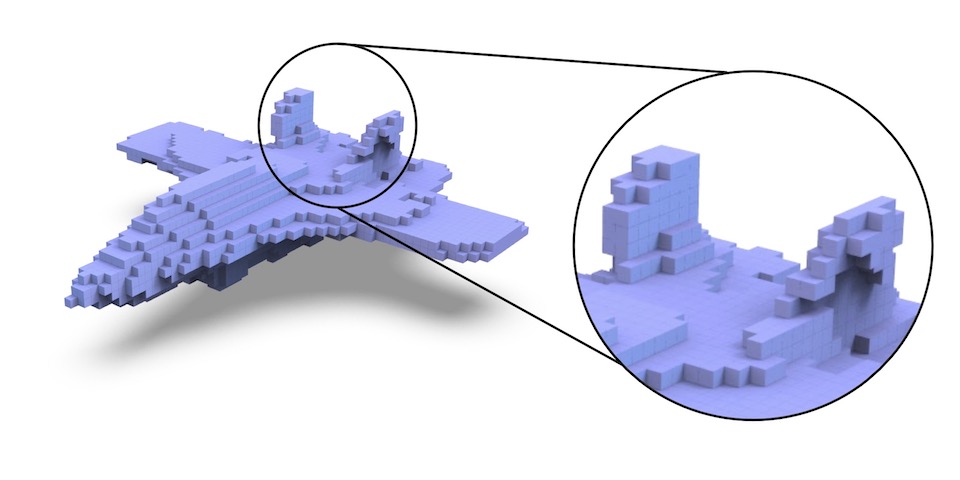} &
\includegraphics[trim={3cm 3cm 3cm 3cm},clip,width=1.00\linewidth]{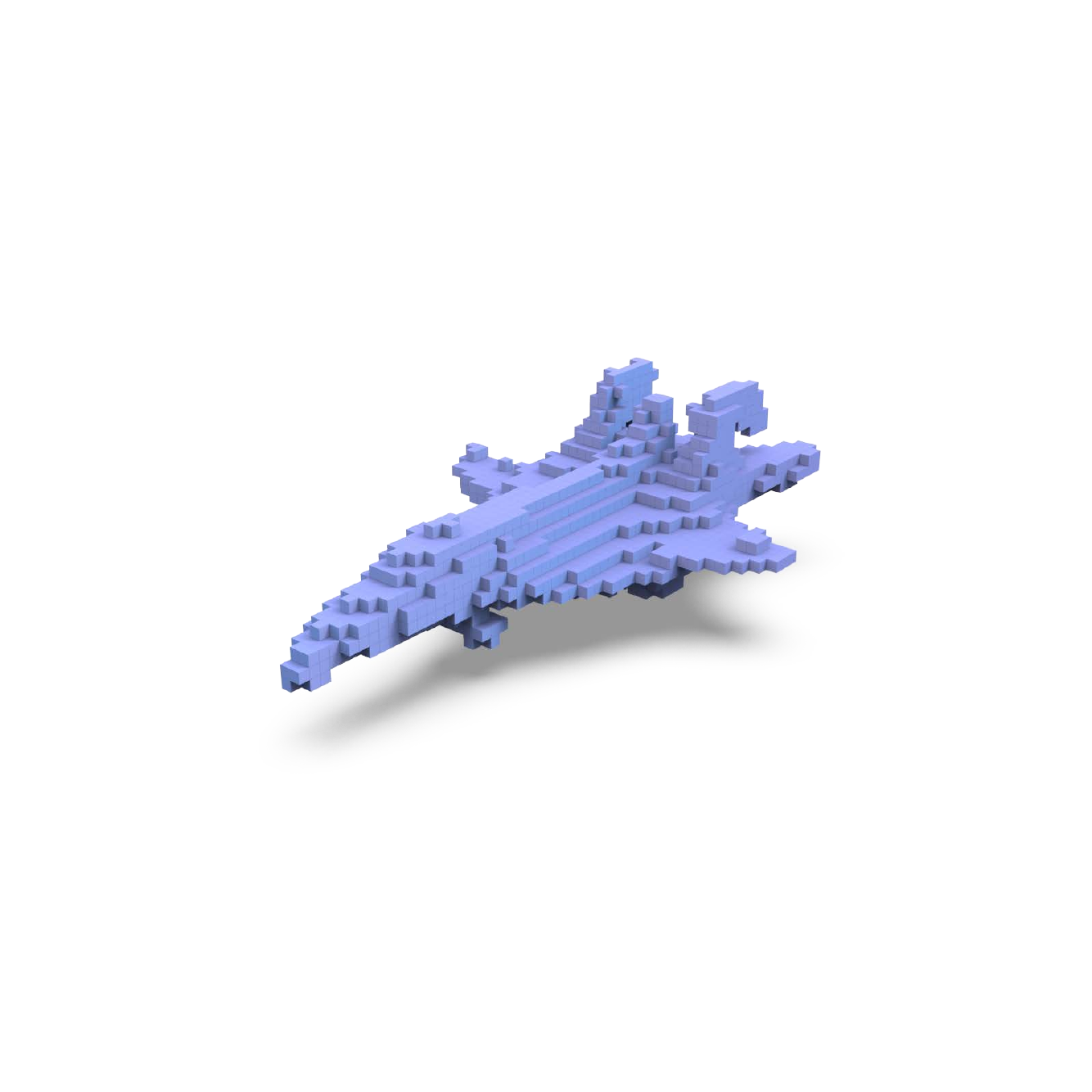} &
\includegraphics[trim={3cm 3cm 3cm 3cm},clip,width=1.00\linewidth]{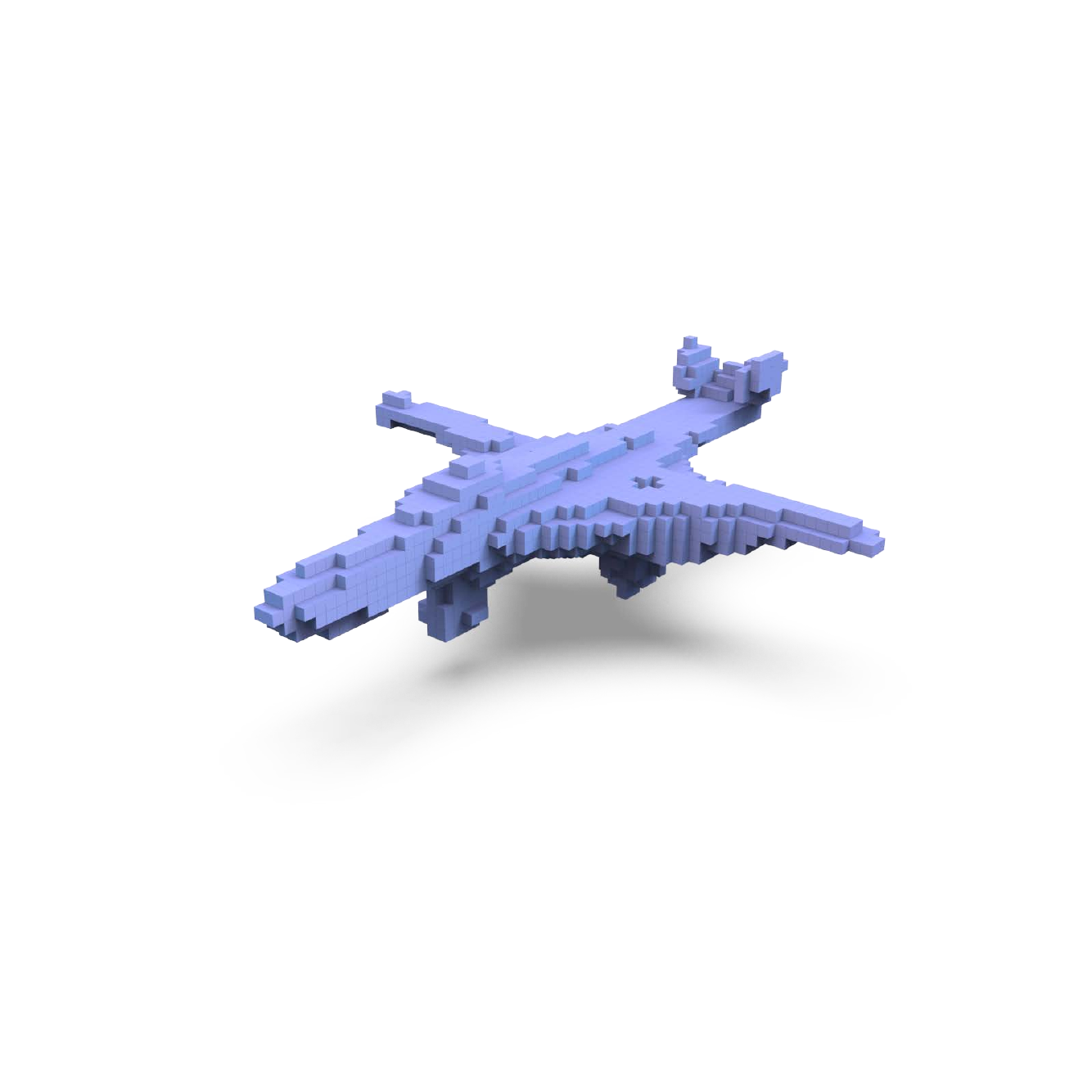} &
\includegraphics[width=1.00\linewidth]{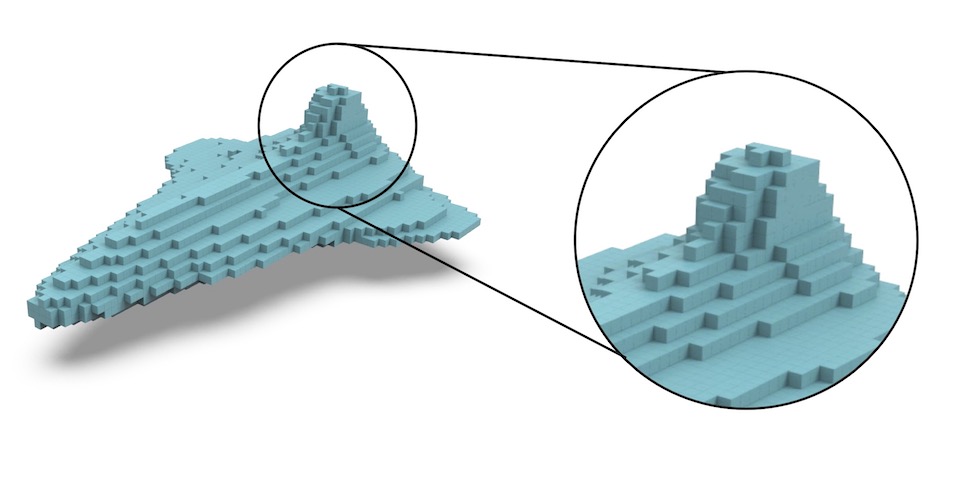} &
\includegraphics[trim={2cm 2cm 2cm 2cm},clip,width=1.00\linewidth]{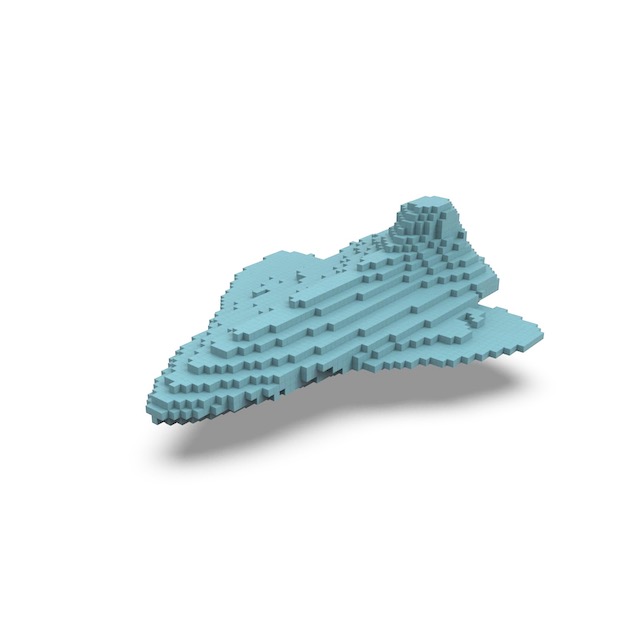} &
\includegraphics[trim={2cm 2cm 2cm 2cm},clip,width=1.00\linewidth]{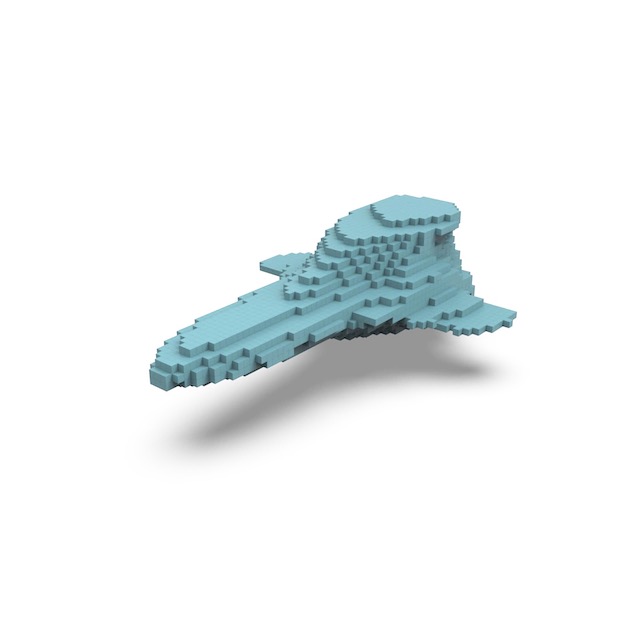} &
\includegraphics[trim={2.5cm 2.5cm 2.5cm 2.5cm},clip,width=1.00\linewidth]{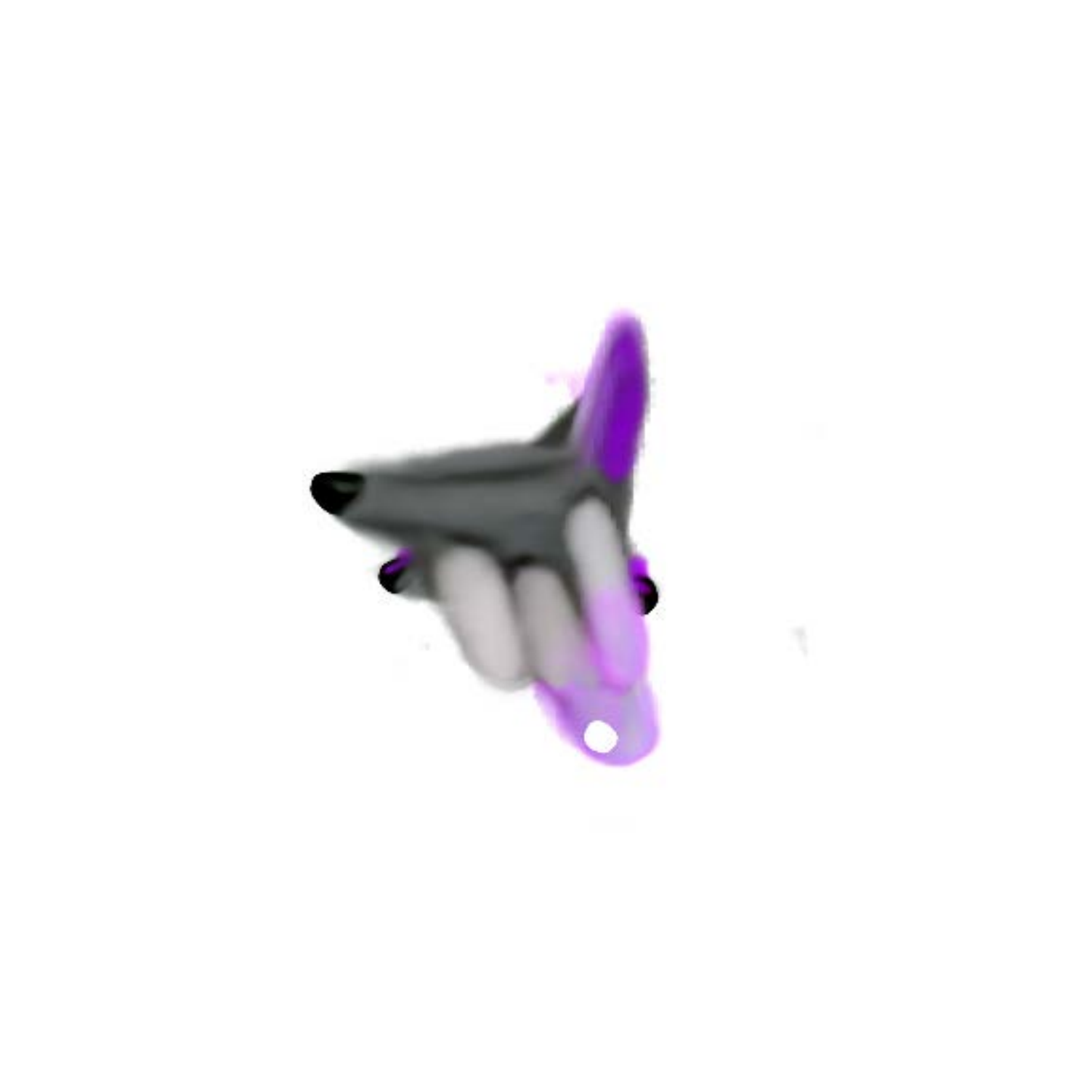} &
\includegraphics[trim={3cm 3cm 3cm 3cm },clip,width=1.00\linewidth]{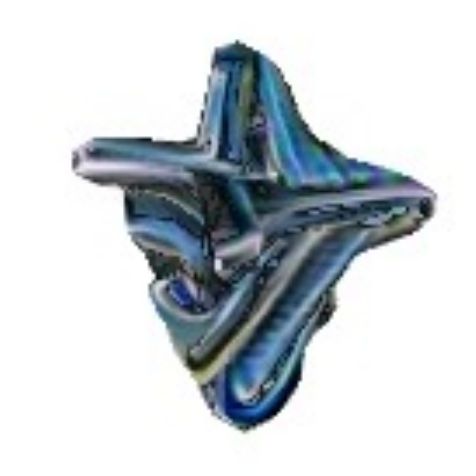}\\
\multicolumn{8}{c}{\small{\texttt{"a jet"}}}
\\

\includegraphics[width=1.00\linewidth]{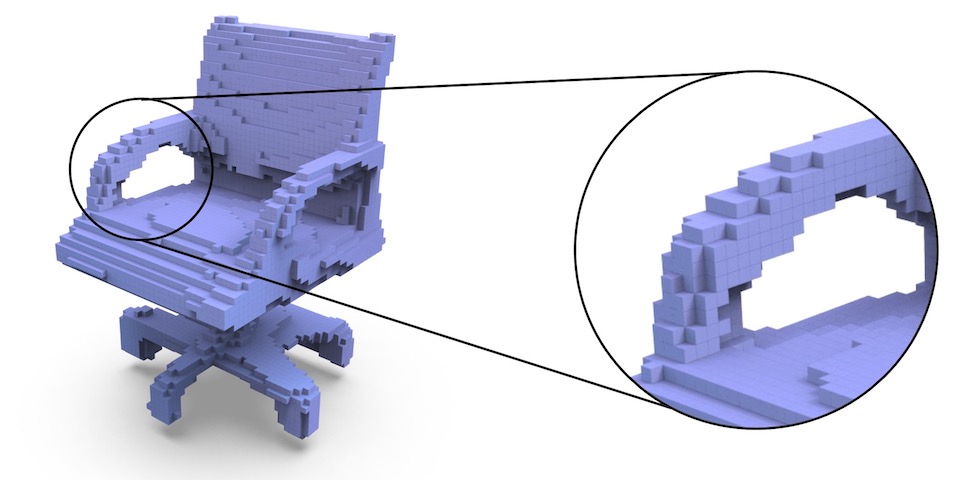} &
\includegraphics[trim={0.5cm 0.5cm 0.5cm 1cm},clip,width=1.00\linewidth]{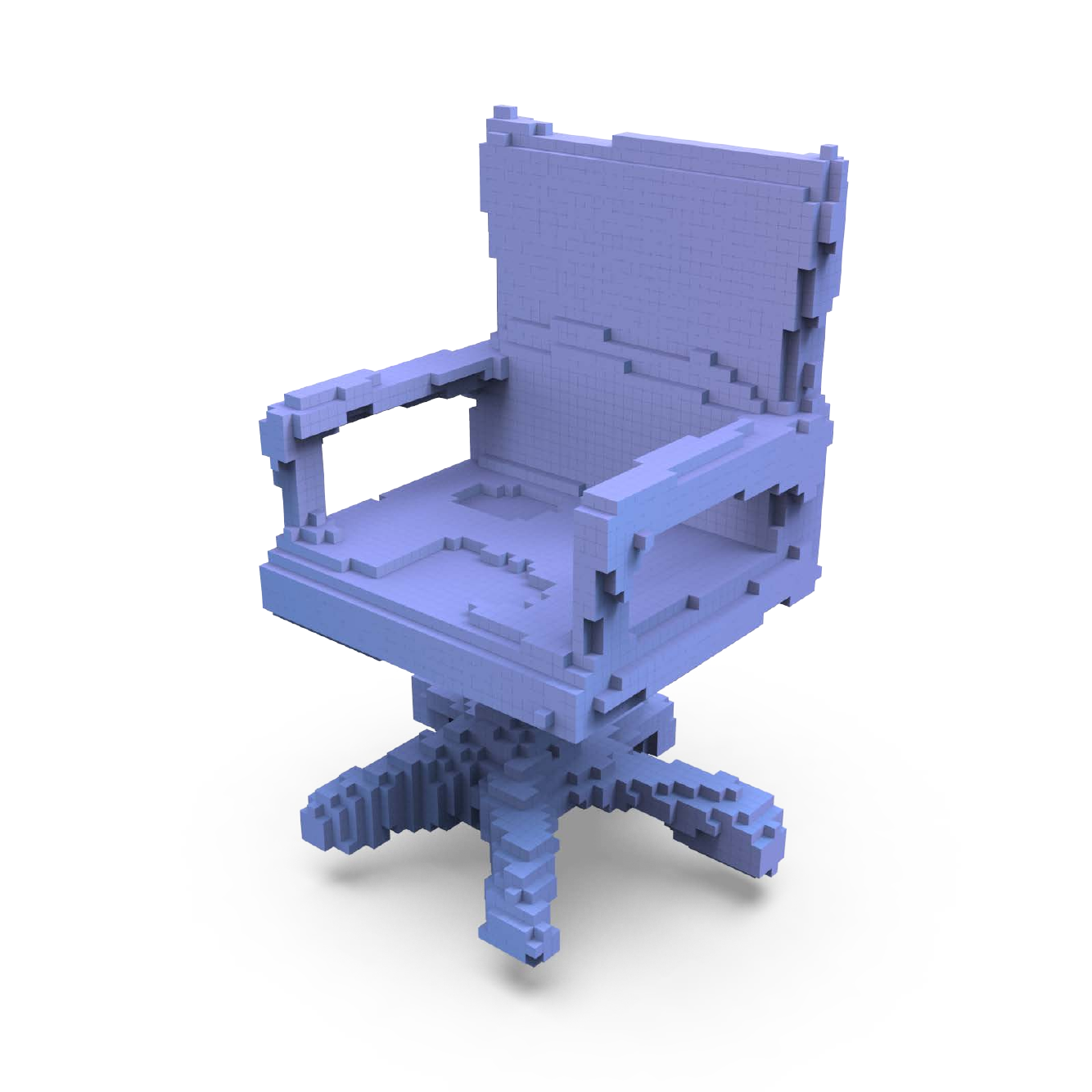} &
\includegraphics[trim={0.5cm 0.5cm 0.5cm 1cm},clip,width=1.00\linewidth]{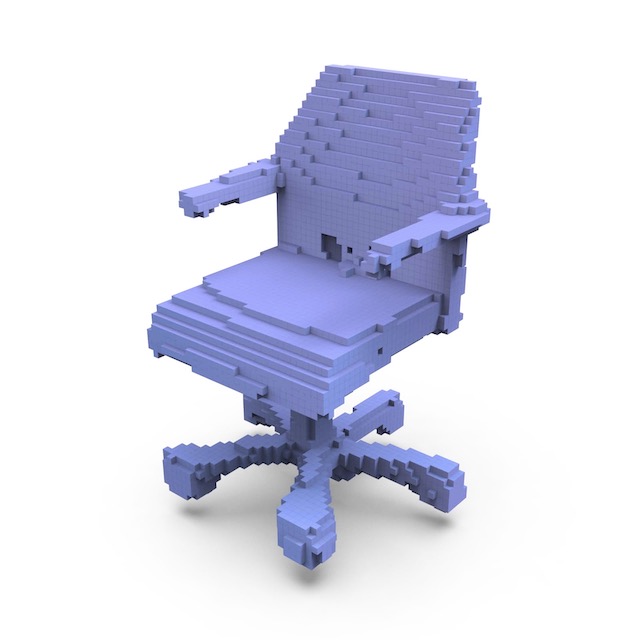} &
\includegraphics[width=1.00\linewidth]{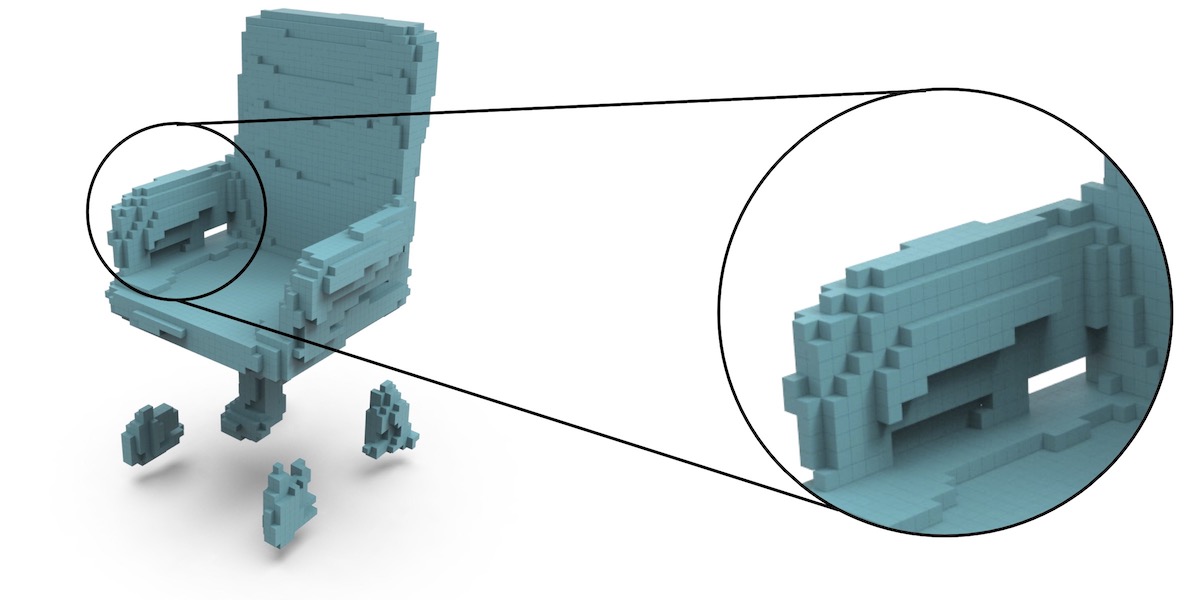} &
\includegraphics[width=1.00\linewidth]{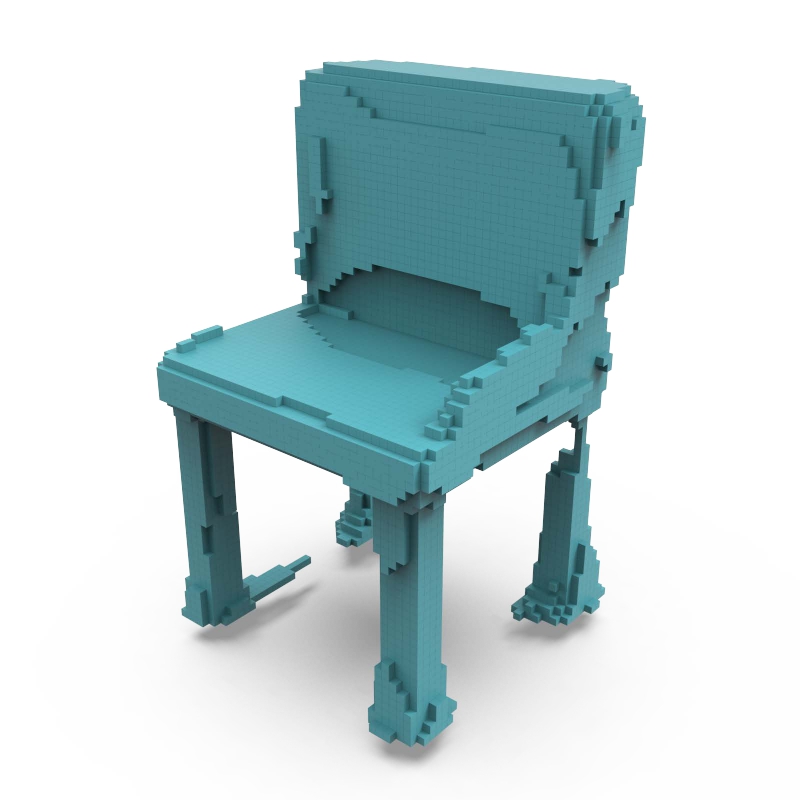} &
\includegraphics[width=1.00\linewidth]{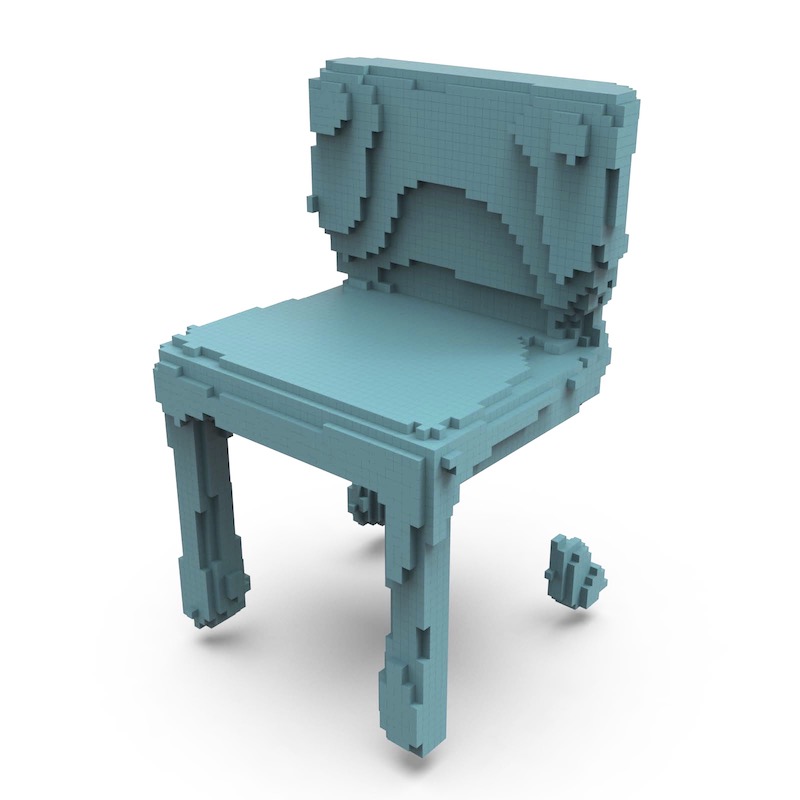} &
\includegraphics[trim={2.5cm 2.5cm 2.5cm 2.5cm},clip,width=1.00\linewidth]{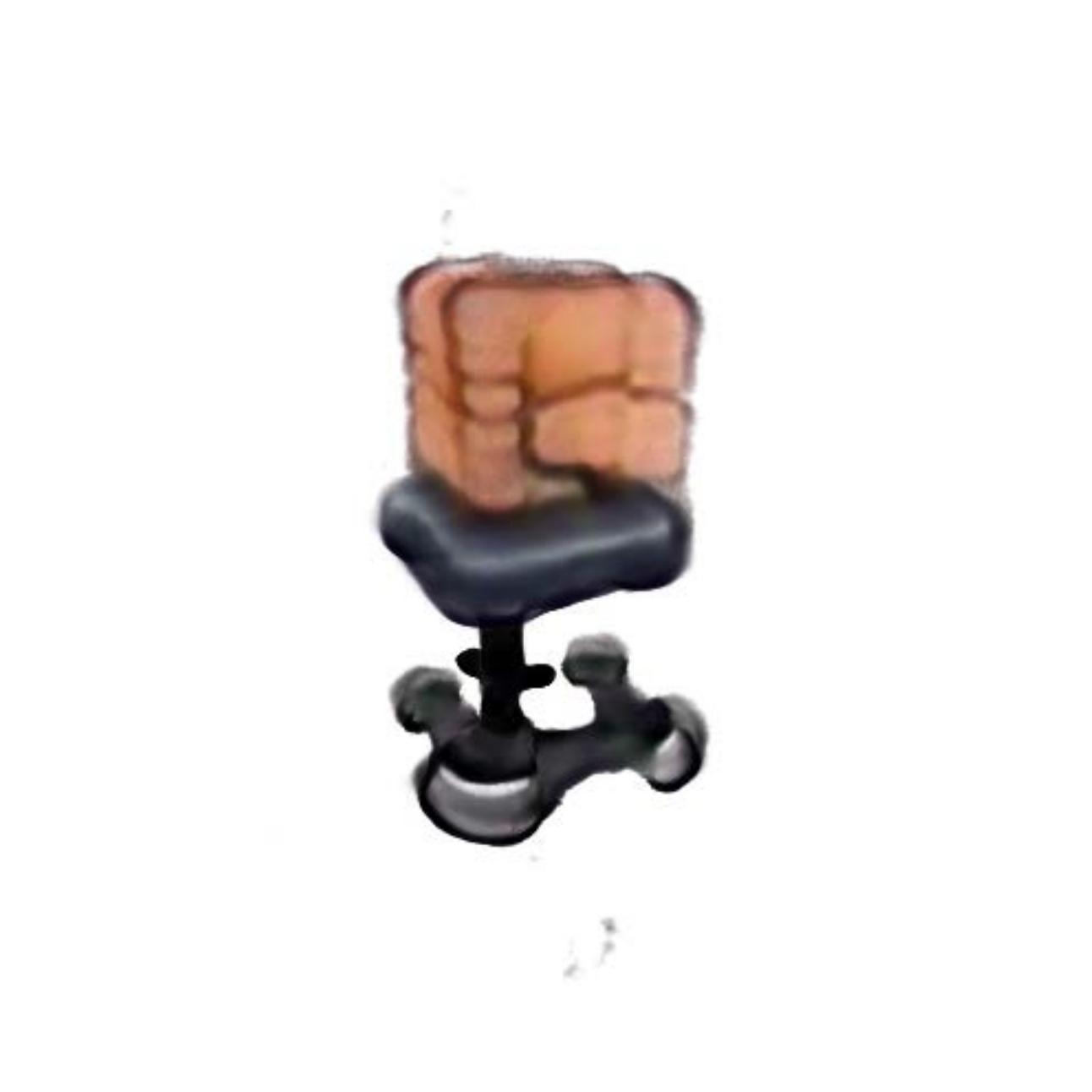} &
\includegraphics[trim={3cm 3cm 3cm 3cm },clip,width=1.00\linewidth]{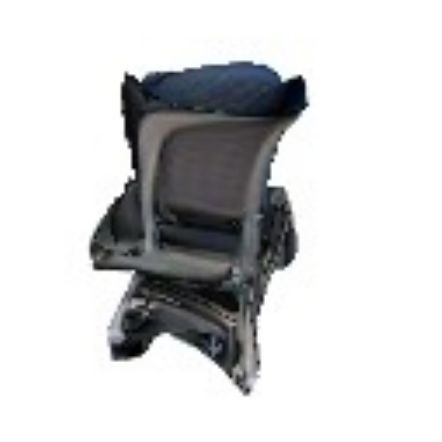}\\
\multicolumn{8}{c}{\small{\texttt{"an office chair"}}}
\\

\includegraphics[width=1.00\linewidth]{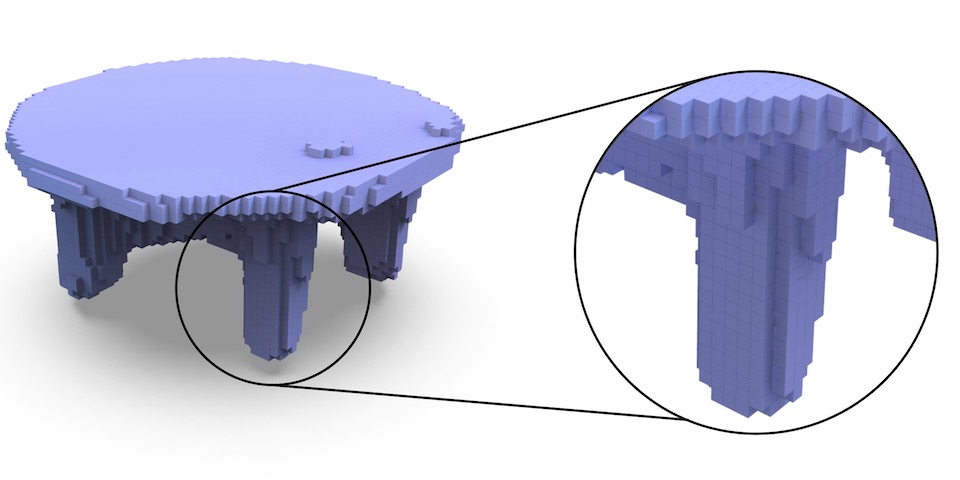} &
\includegraphics[trim={0.5cm 0.5cm 0.5cm 1cm},clip,width=1.00\linewidth]{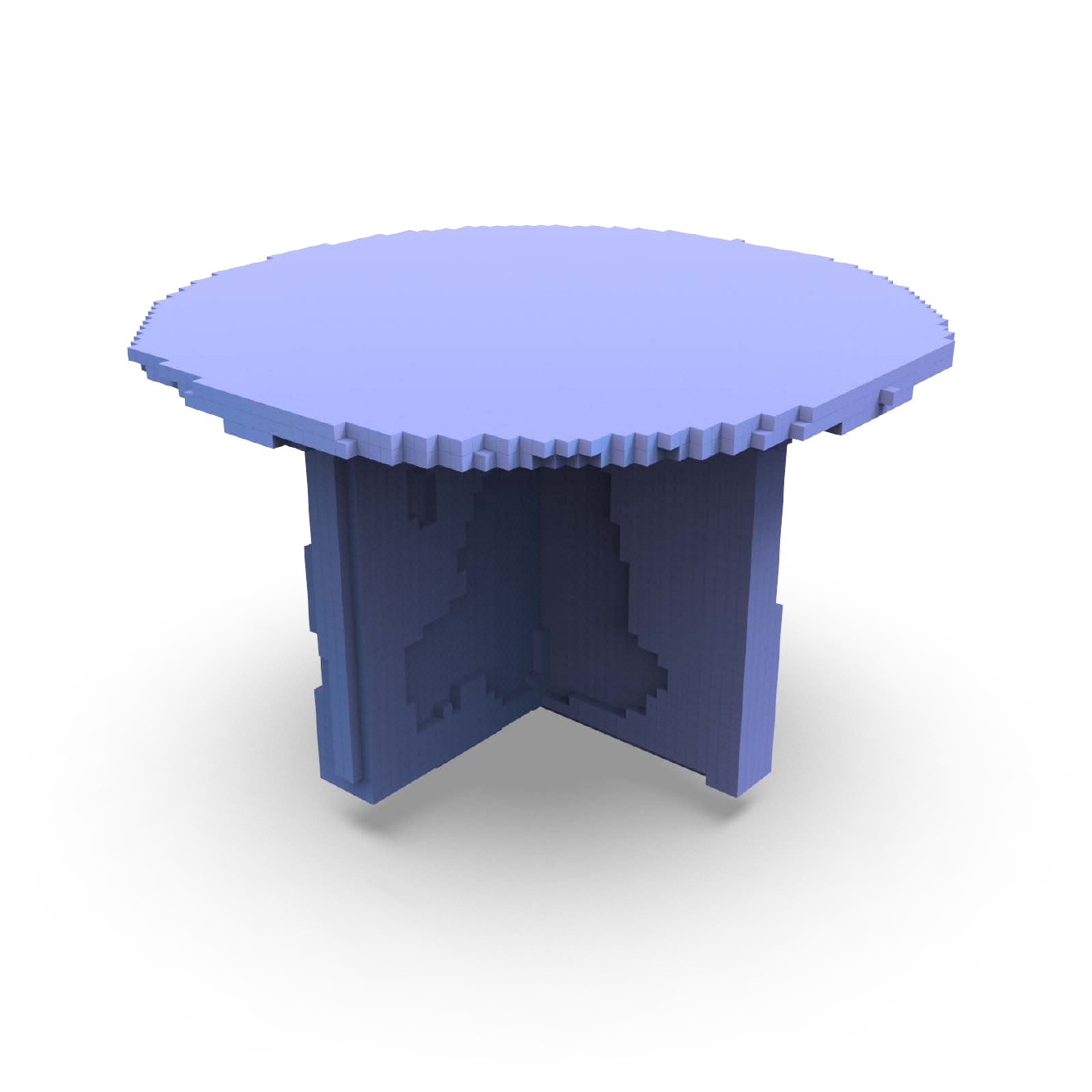} &
\includegraphics[trim={0.5cm 0.5cm 0.5cm 1cm},clip,width=1.00\linewidth]{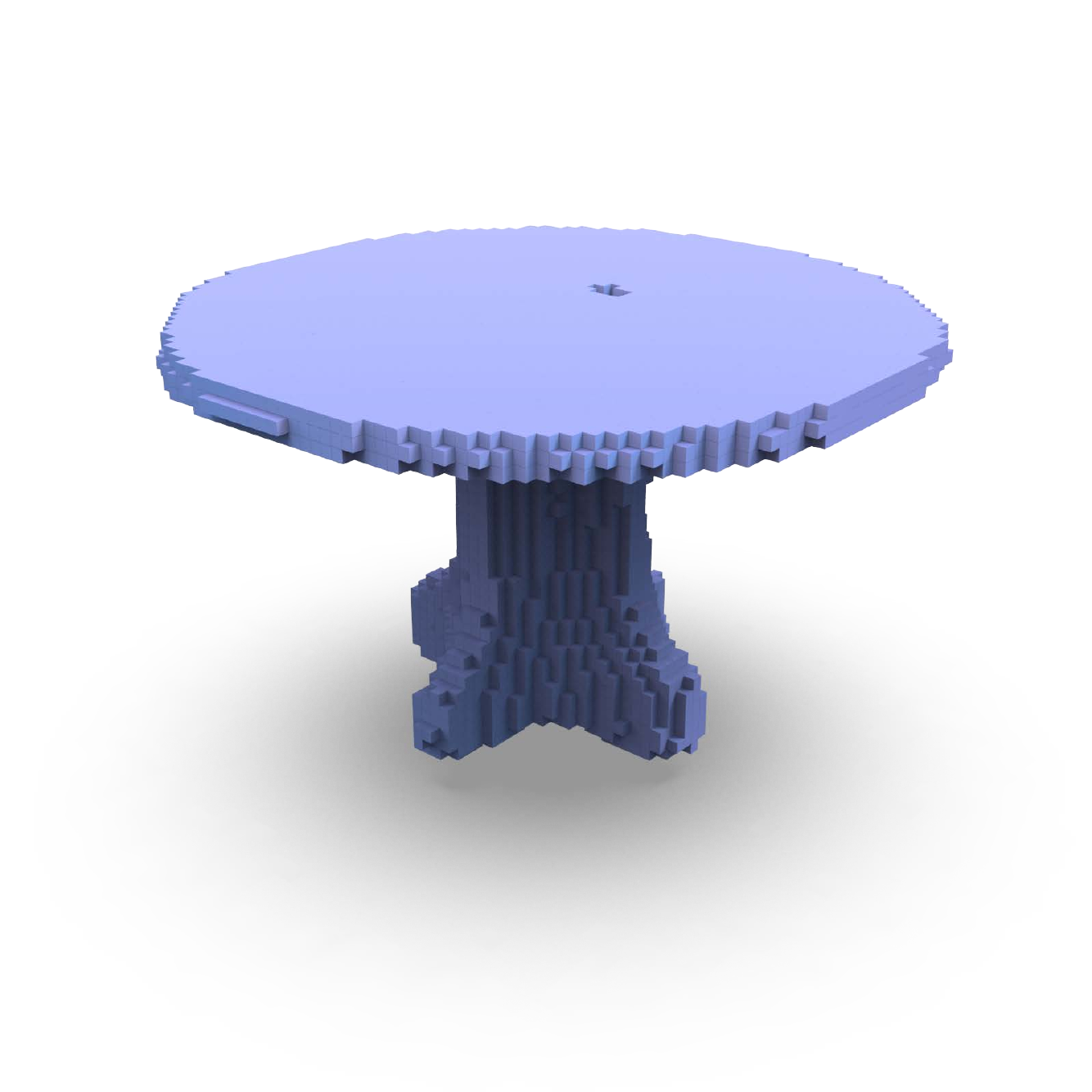} &
\includegraphics[width=1.00\linewidth]{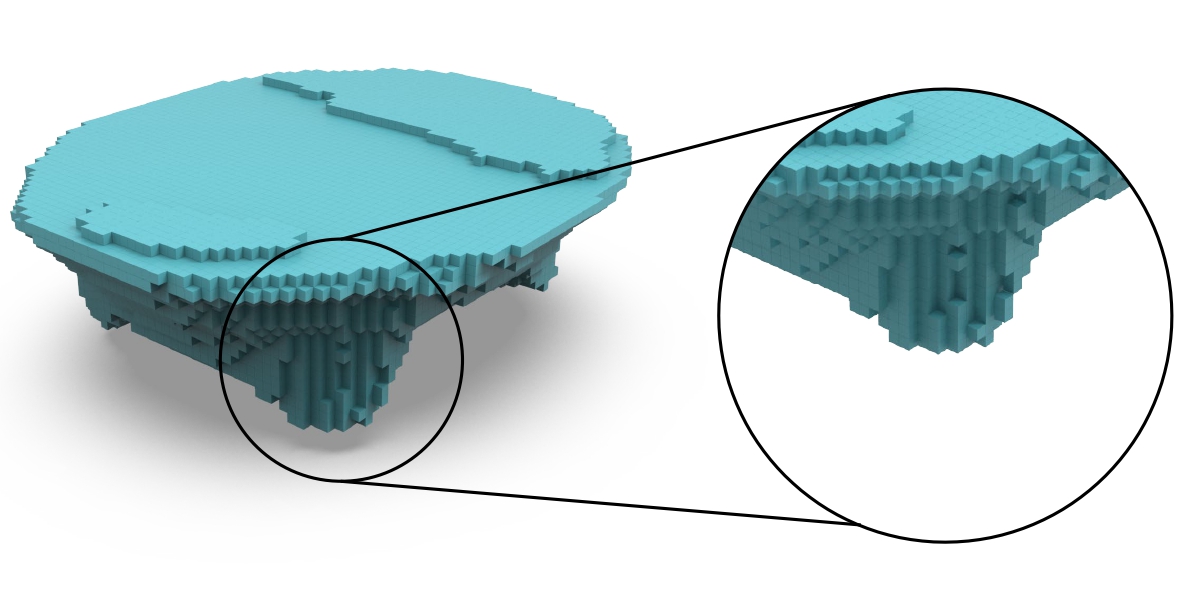} &
\includegraphics[trim={0.5cm 0.5cm 0.5cm 1cm},clip,width=1.00\linewidth]{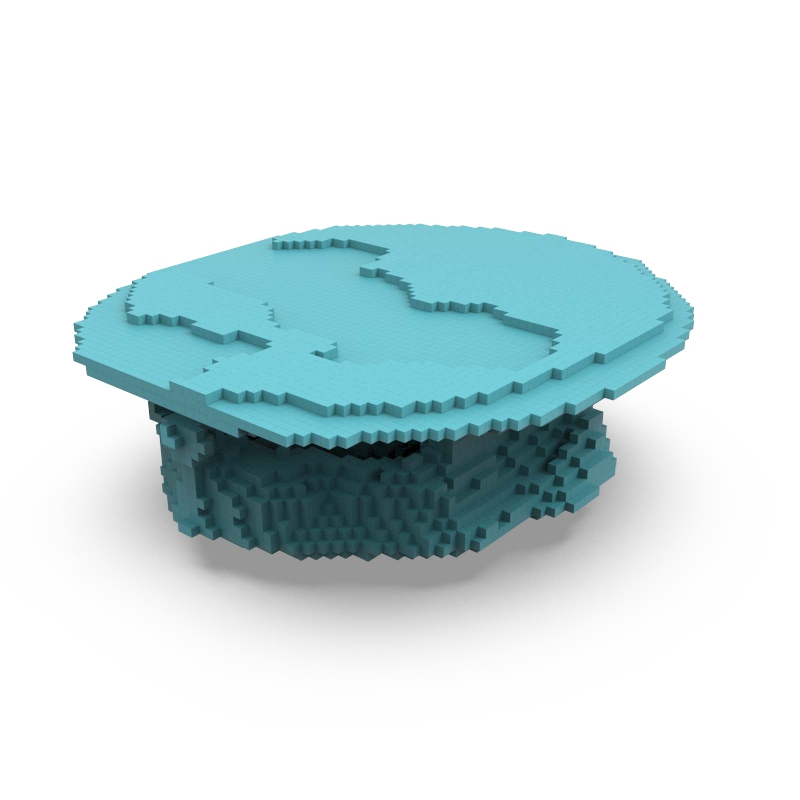} &
\includegraphics[trim={0.5cm 0.5cm 0.5cm 1cm},clip,width=1.00\linewidth]{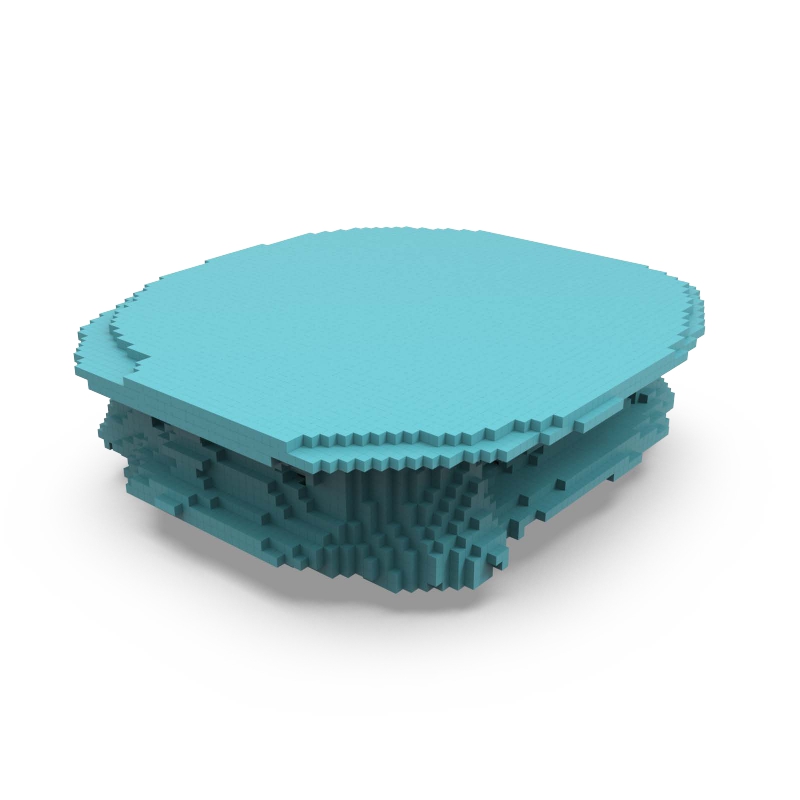} &
\includegraphics[trim={2.5cm 2.5cm 2.5cm 2.5cm},clip,width=1.00\linewidth]{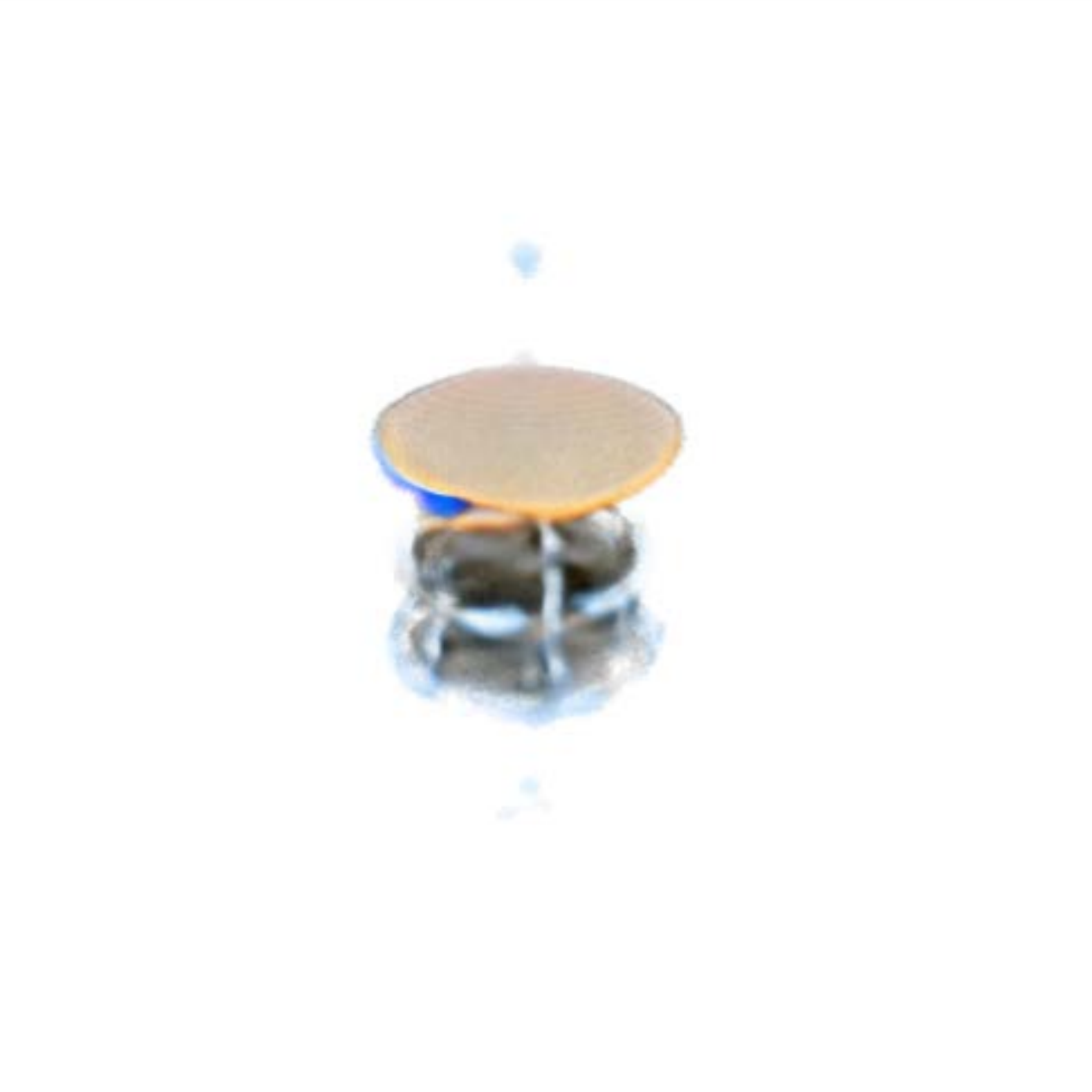} &
\includegraphics[trim={3cm 3cm 3cm 3cm },clip,width=1.00\linewidth]{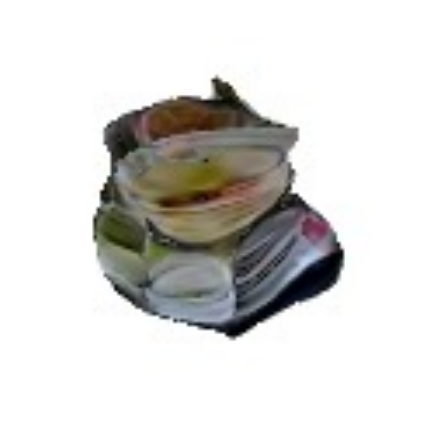}\\
\multicolumn{8}{c}{\small{\texttt{"a round table"}}}
\\

\includegraphics[width=1.00\linewidth]{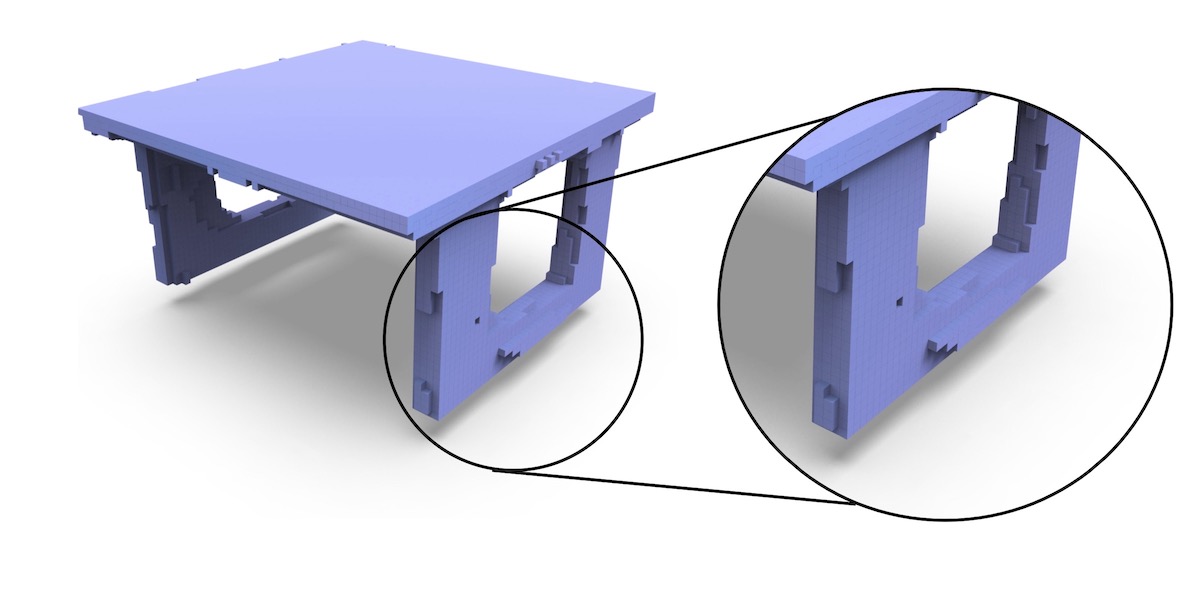} &
\includegraphics[trim={0.5cm 0.5cm 0.5cm 1cm},clip,width=1.00\linewidth]{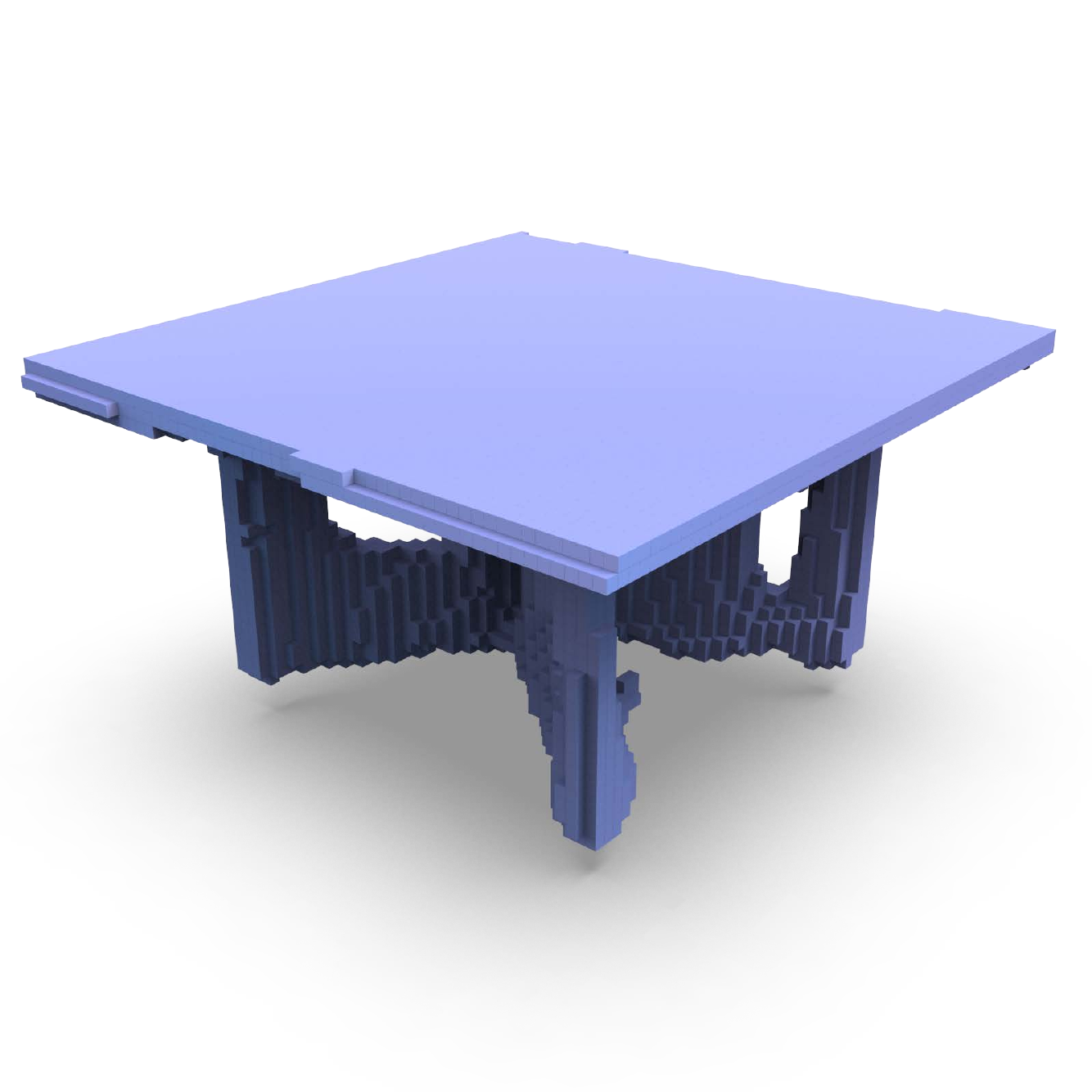} &
\includegraphics[trim={0.5cm 0.5cm 0.5cm 1cm},clip,width=1.00\linewidth]{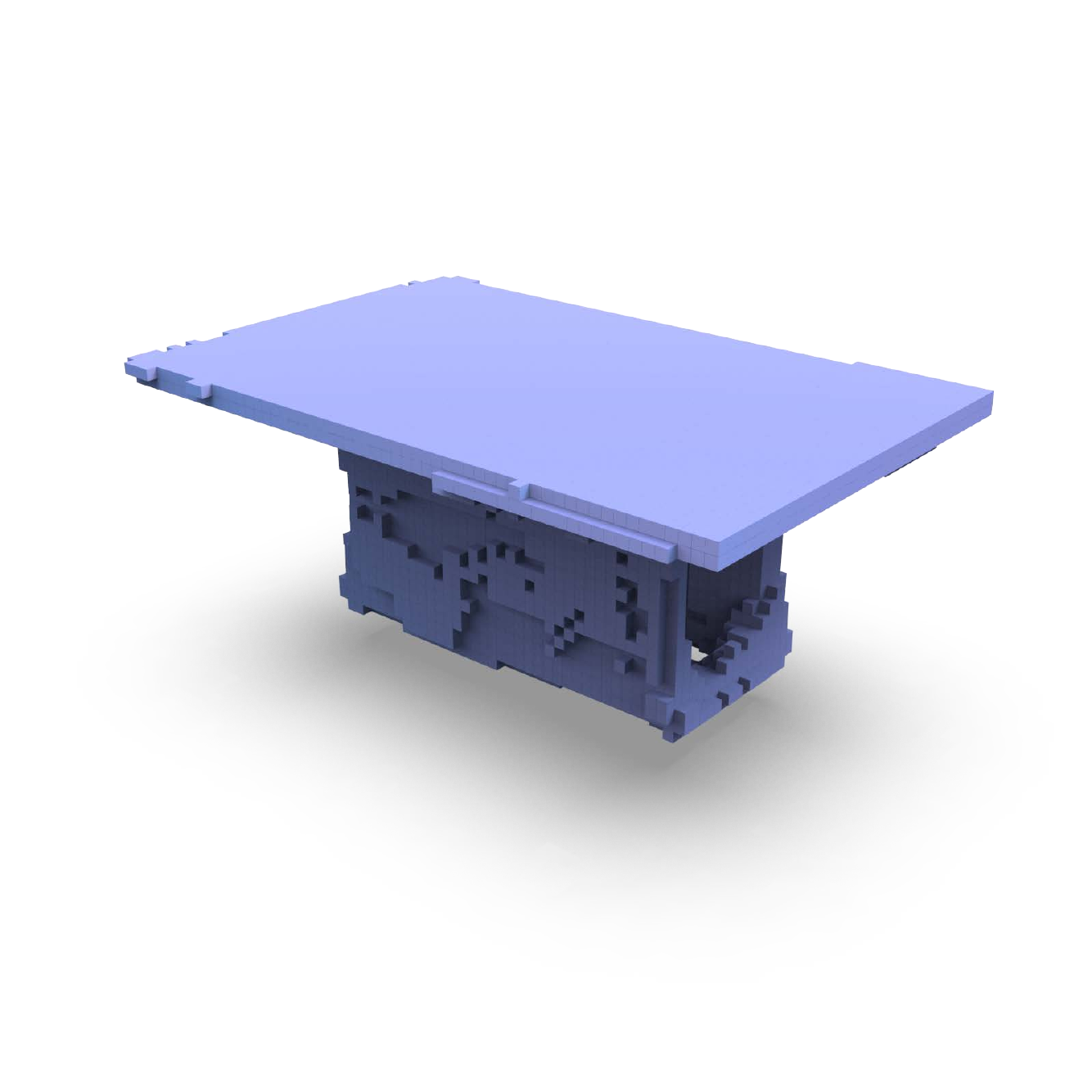} &
\includegraphics[width=1.00\linewidth]{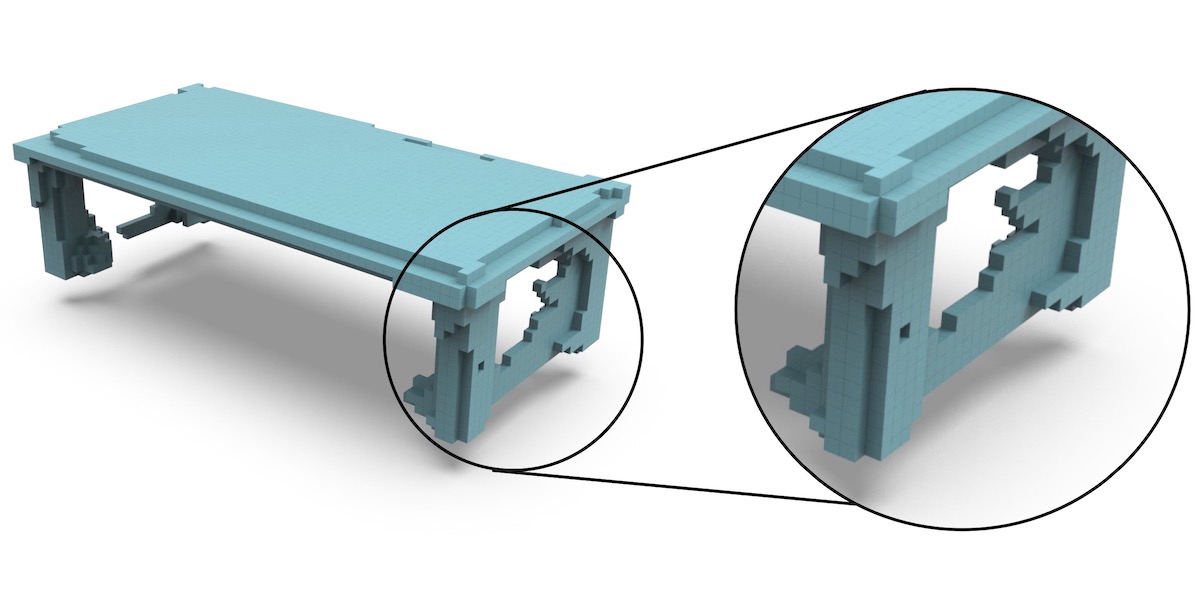} &
\includegraphics[trim={0.5cm 0.5cm 0.5cm 1cm},clip,width=1.00\linewidth]{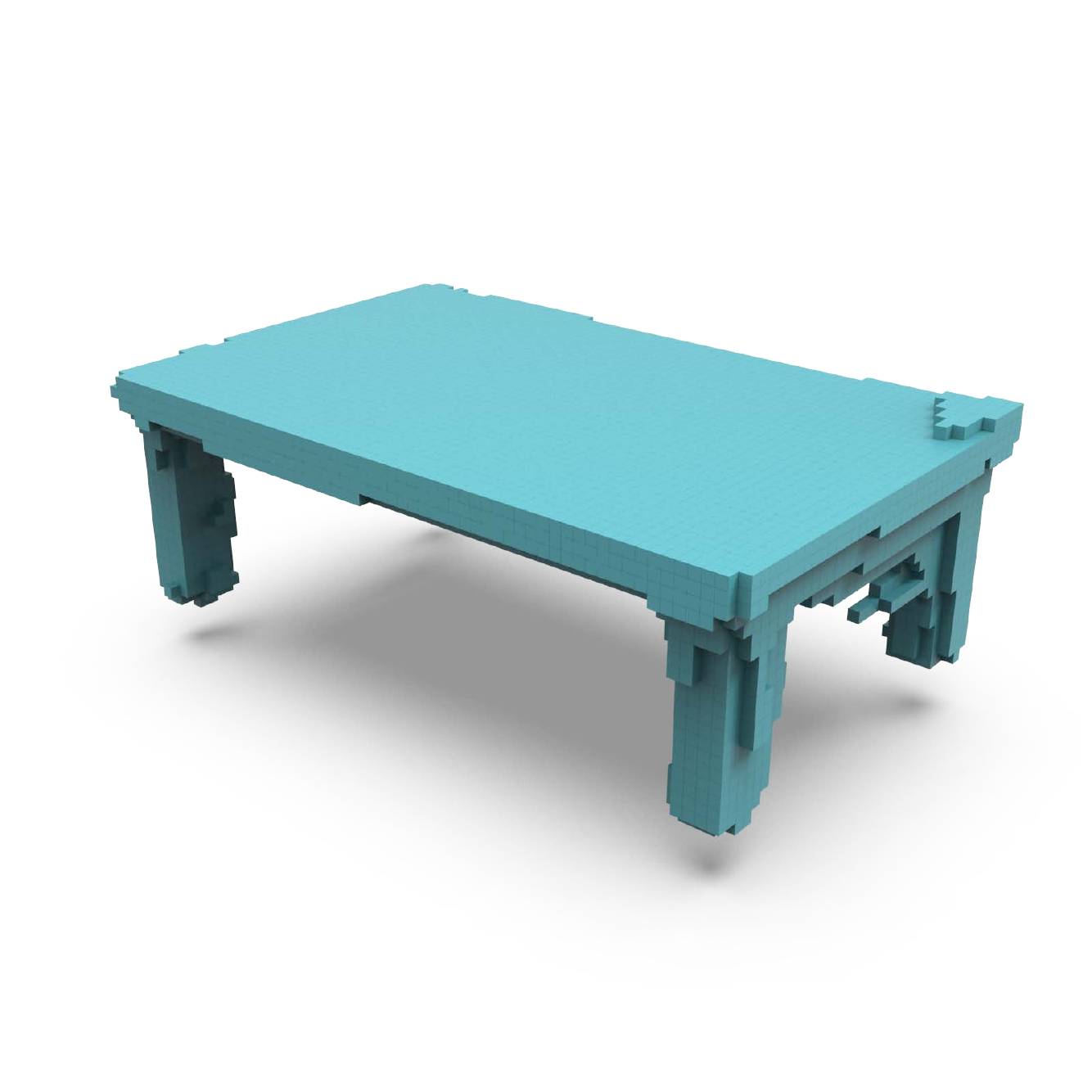} &
\includegraphics[trim={0.5cm 0.5cm 0.5cm 1cm},clip,width=1.00\linewidth]{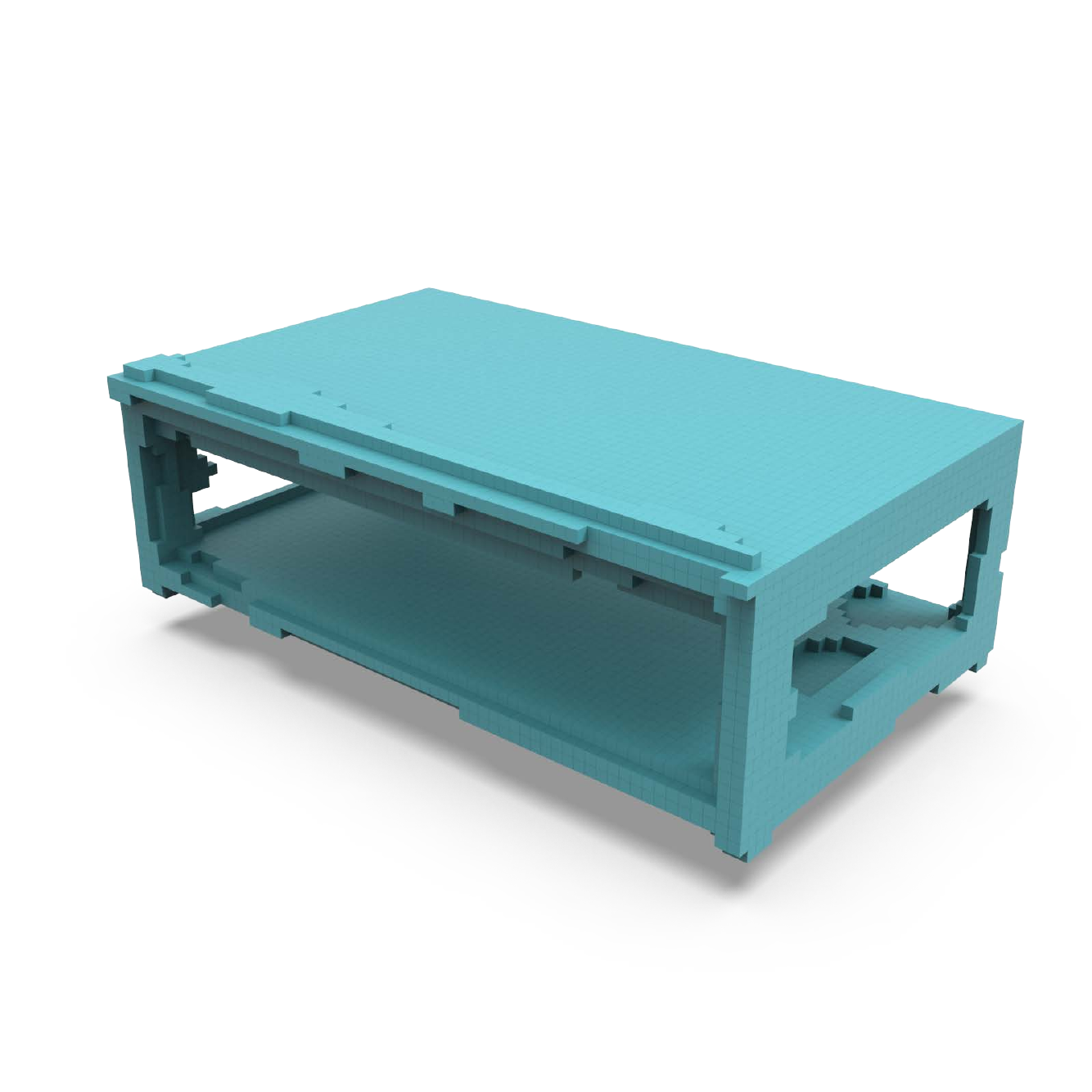} &
\includegraphics[trim={2.5cm 2.5cm 2.5cm 1cm},clip,width=1.00\linewidth]{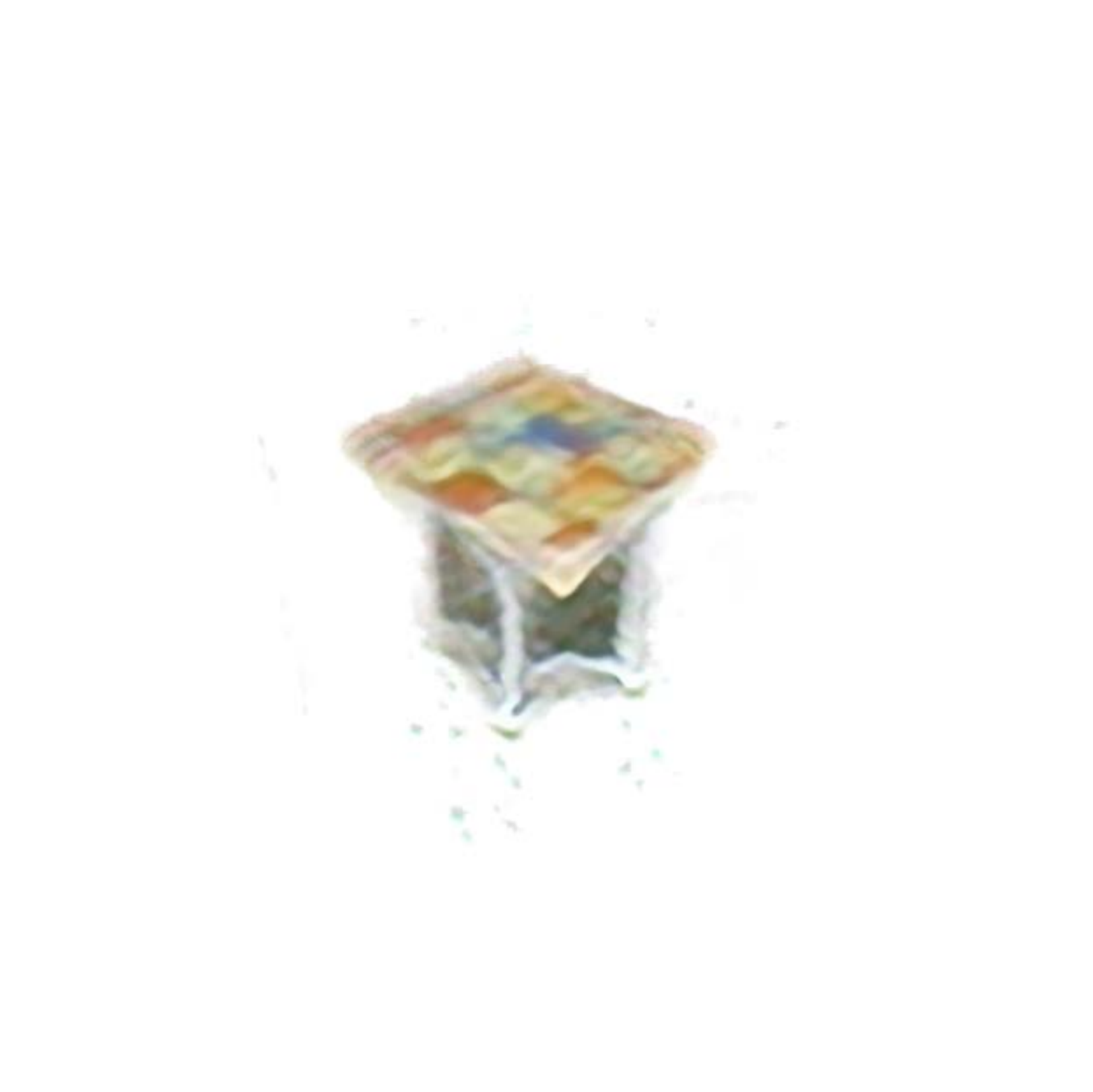} &
\includegraphics[trim={3cm 3cm 3cm 3cm },clip,width=1.00\linewidth]{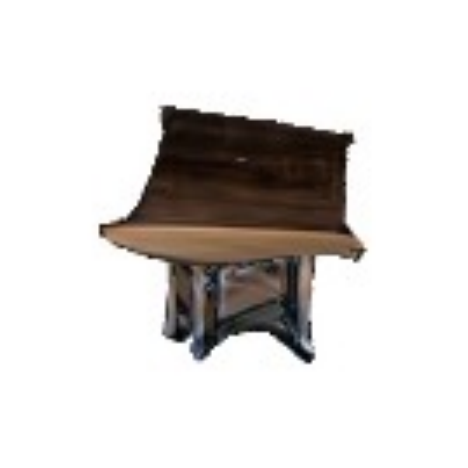}\\
\multicolumn{8}{c}{\small{\texttt{"a rectangular table"}}}
\\
\includegraphics[width=1.00\linewidth]{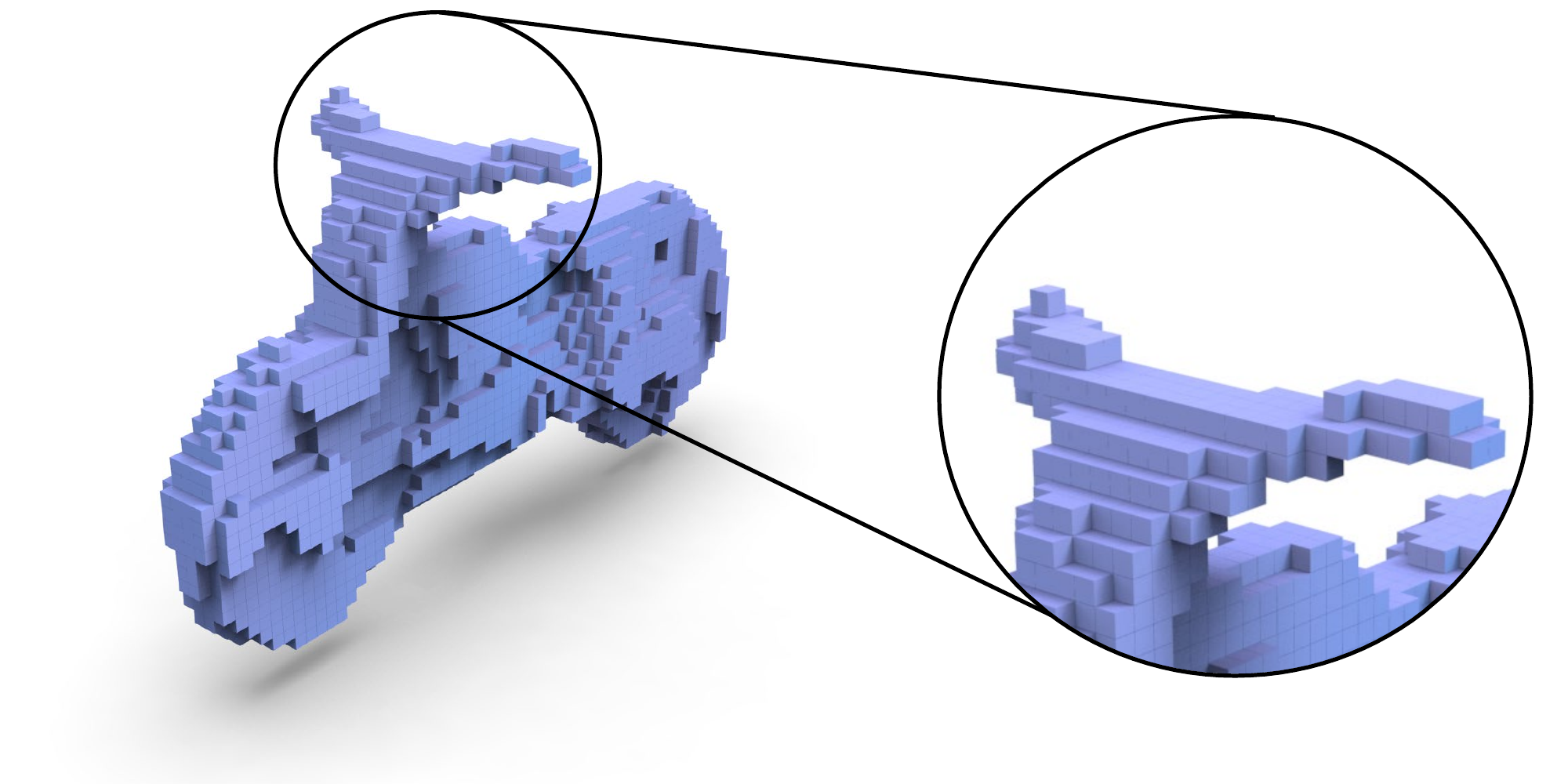} &
\includegraphics[trim={2cm 2cm 2cm 2cm},clip,width=1.00\linewidth]{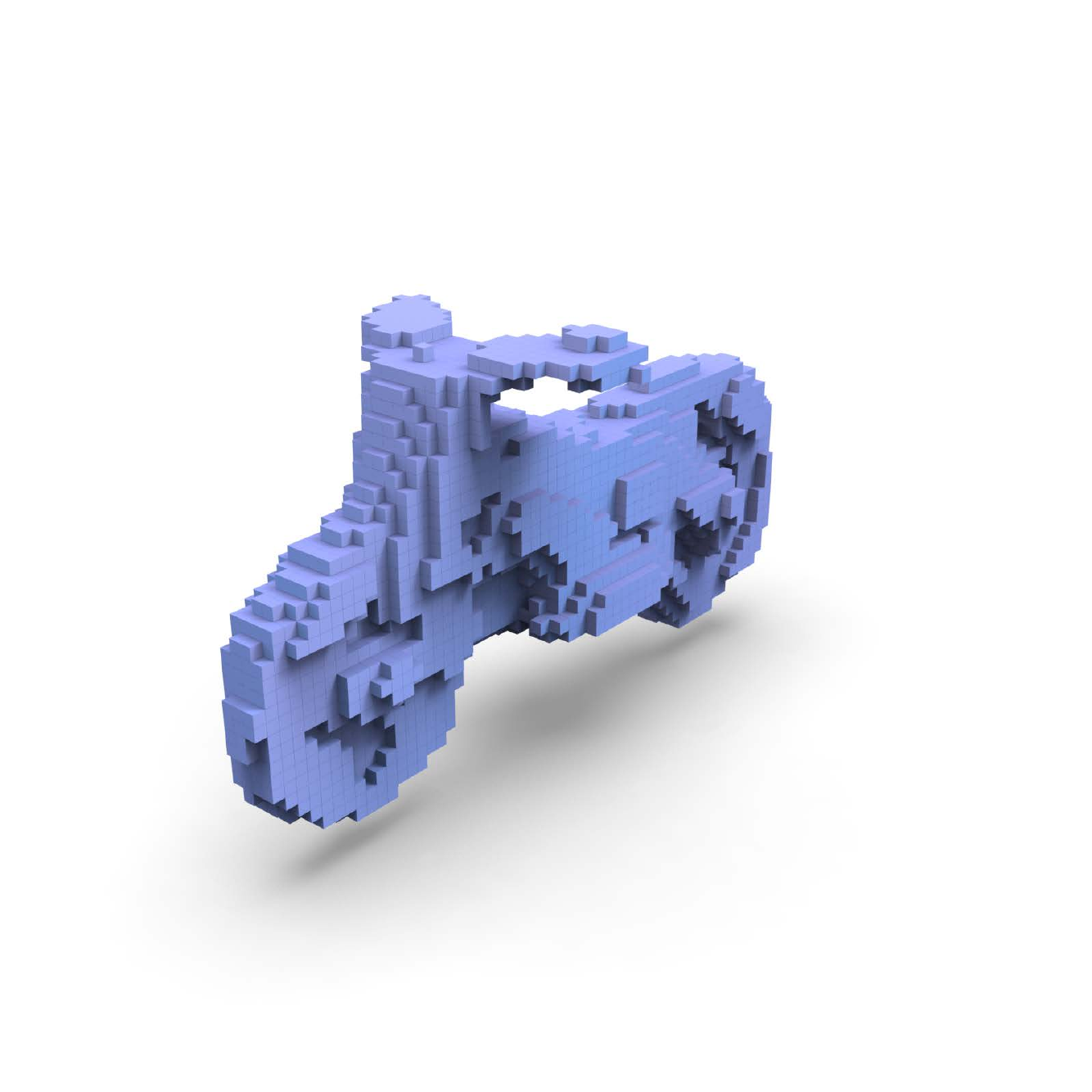} &
\includegraphics[trim={2cm 2cm 2cm 2cm},clip,width=1.00\linewidth]{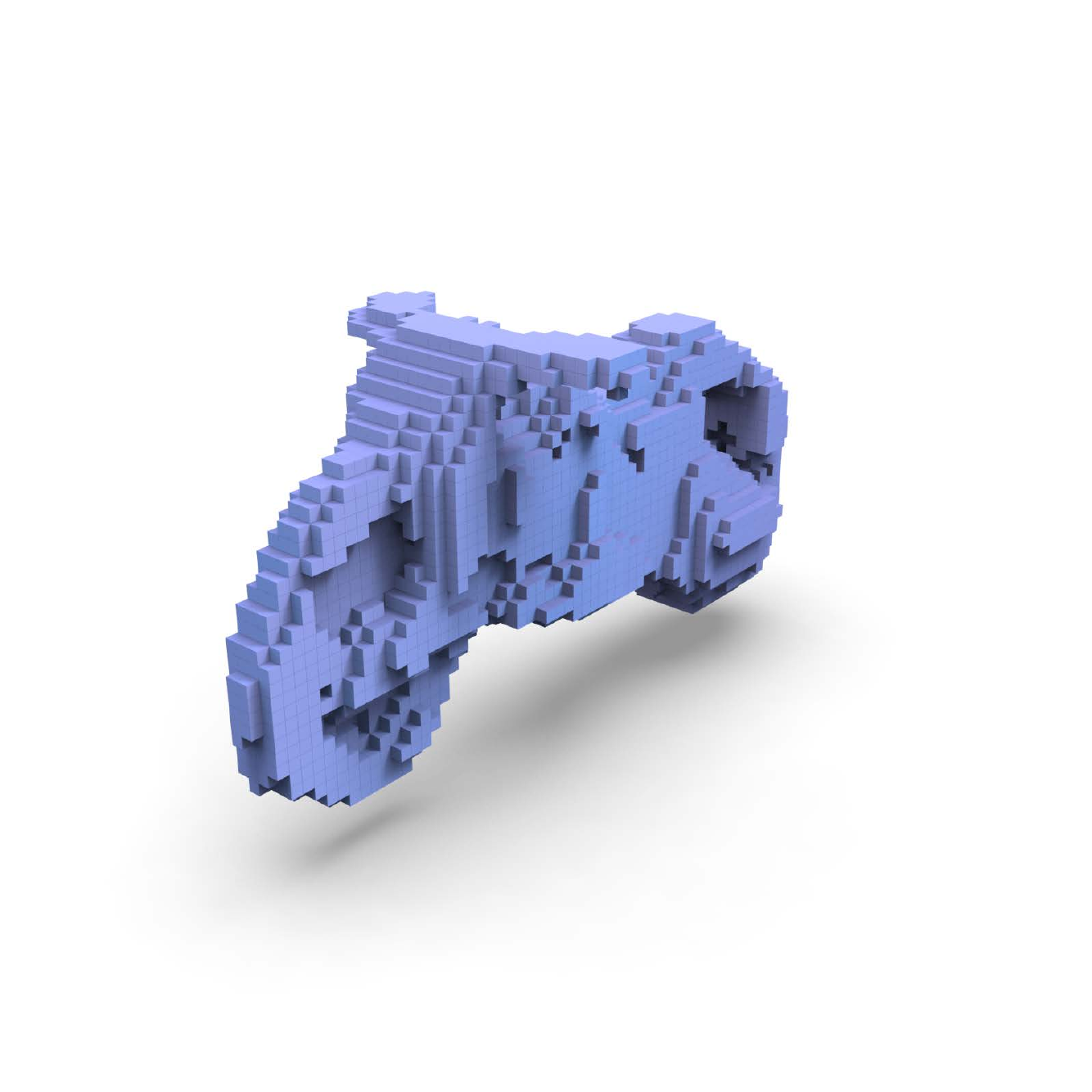} &
\includegraphics[width=1.00\linewidth]{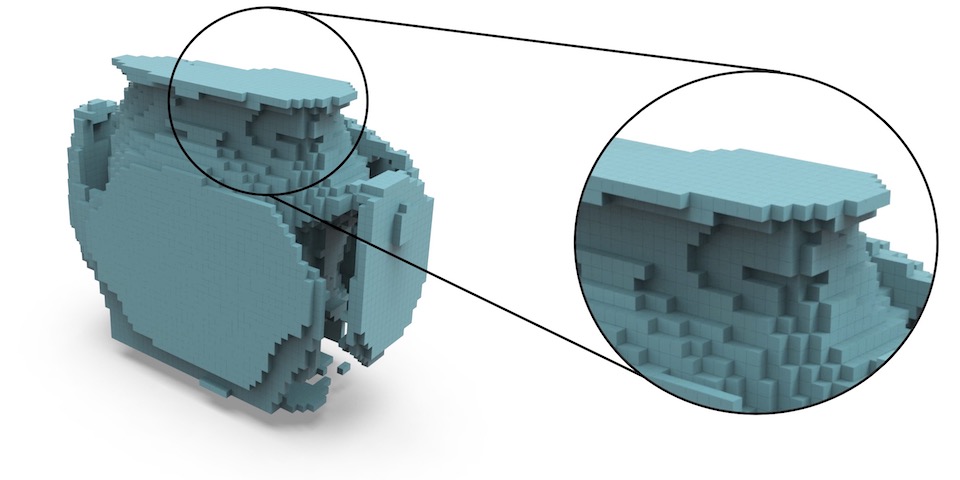} &
\includegraphics[trim={1cm 1cm 1cm 1cm},clip,width=1.00\linewidth]{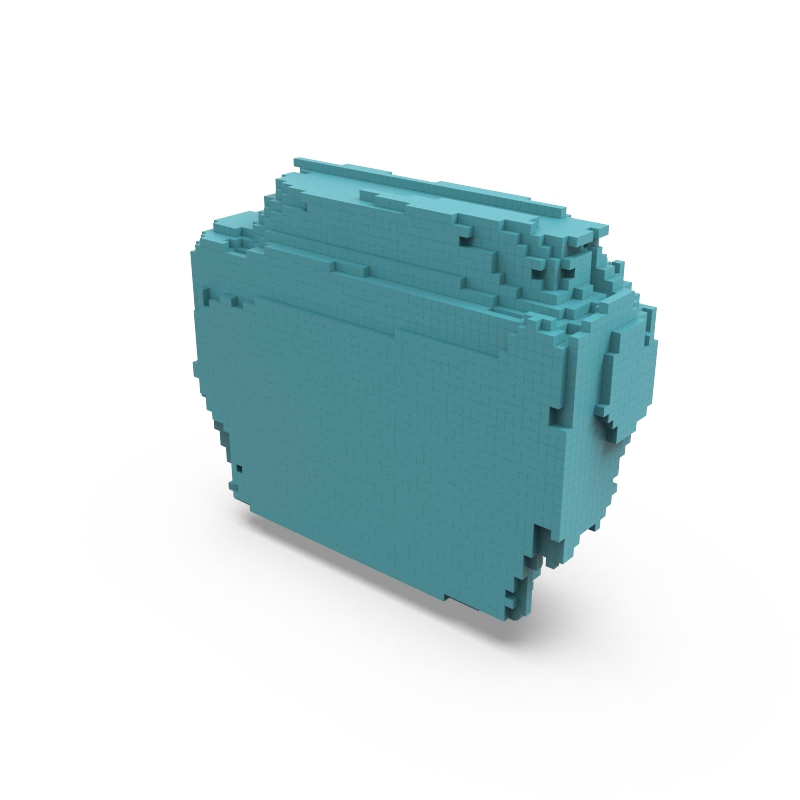} &
\includegraphics[trim={1cm 1cm 1cm 1cm},clip,width=1.00\linewidth]{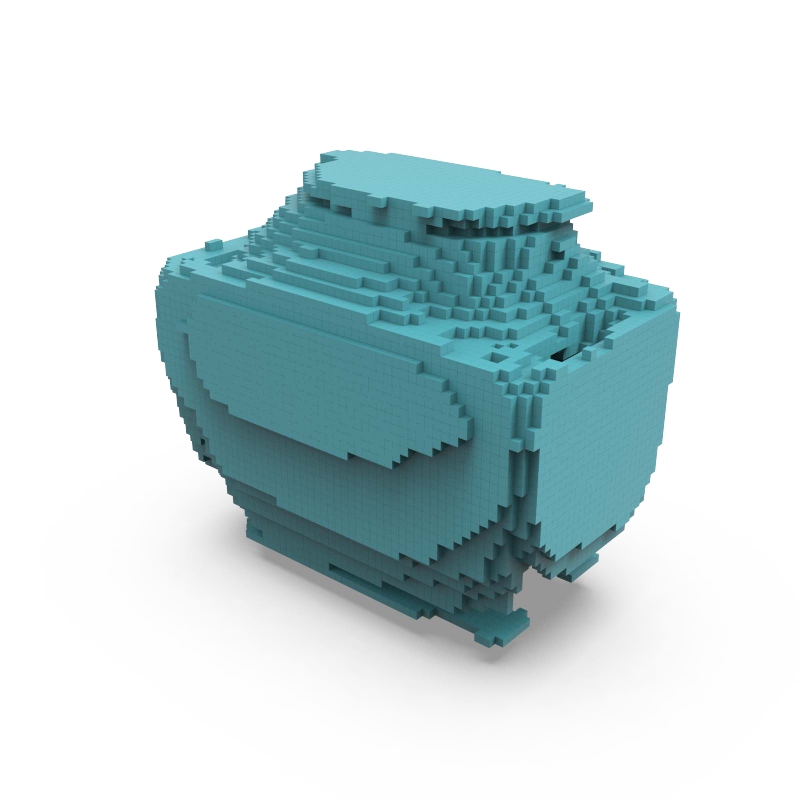} &
\includegraphics[trim={2.5cm 2.5cm 2.5cm 2.5cm},clip,width=1.00\linewidth]{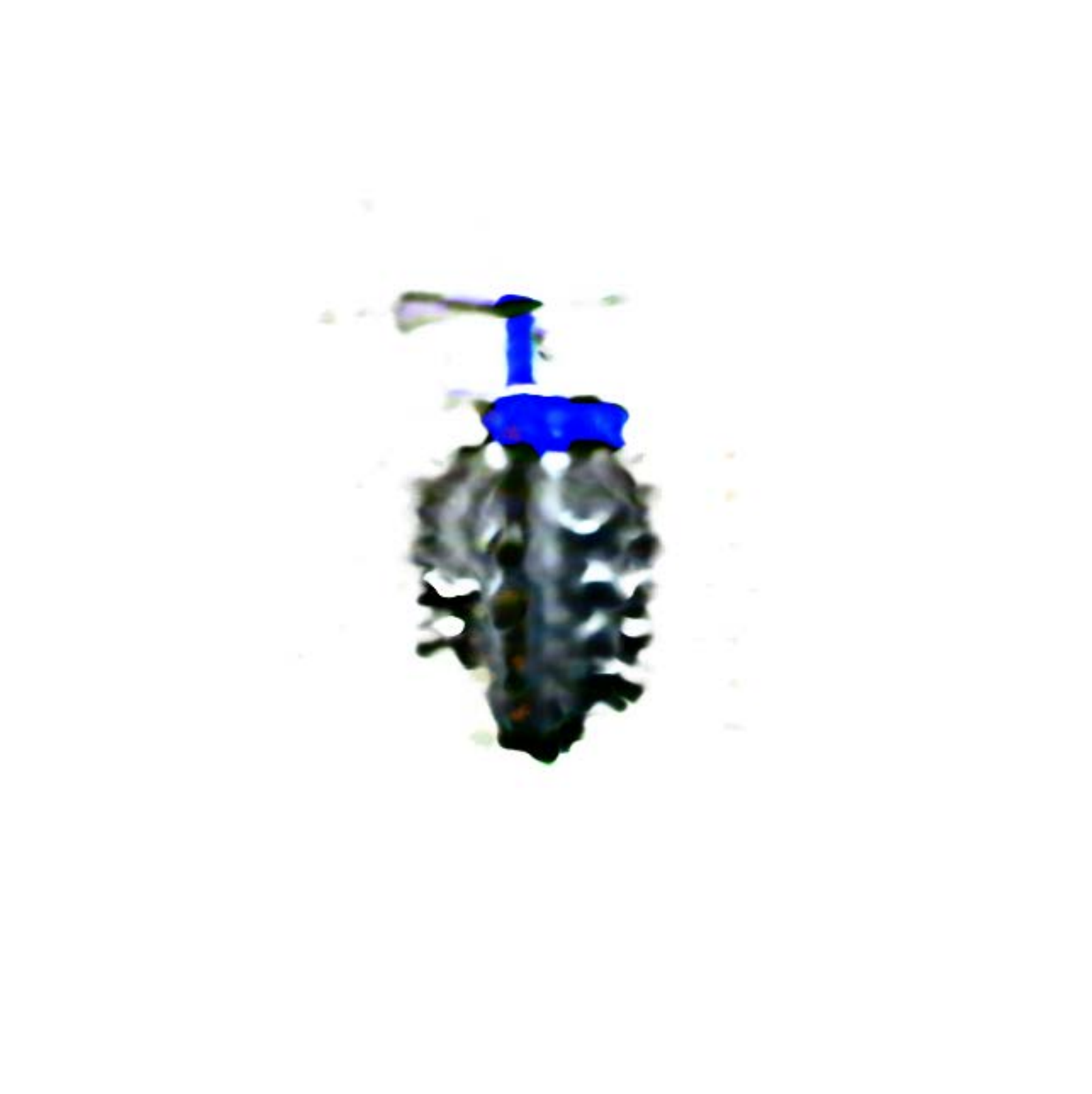} &
\includegraphics[trim={3cm 3cm 3cm 3cm  },clip,width=1.00\linewidth]{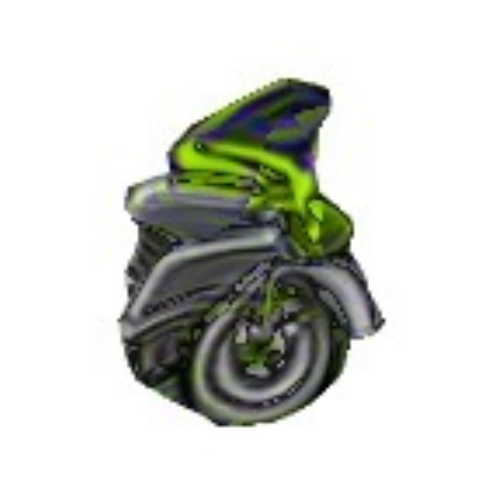}\\
\multicolumn{8}{c}{\small{\texttt{"a motor bike"}}}
\\

\end{tabular}
  \caption{Qualitative comparison of \nickname(rendered in purple), CLIP-Forge~\cite{sanghi2022clip}(rendered in green),  DreamField~\cite{jain2022zero}, and CLIP-Mesh~\cite{khalid2022clip} on text-conditioned generation. For each text prompt, the first generated shape by \nickname and CLIP-Forge~\cite{sanghi2022clip} is zoomed in for detail comparison. Comparing with other methods, \nickname generates a broader category of shapes that correspond with the text prompt. The generated shapes are more diverse and has higher fidelity.
  }
\label{fig:qual_multiple}
\end{figure*}
\begin{table*}
\setlength{\tabcolsep}{7pt}
\caption{Comparisons of CLIP-Forge(CF) baseline (across different sampling strategies), AutoSDF augmented with CLIP (ZS-ASDF) with \nickname(CS) on Accuracy(ACC) and FID. \nickname outperforms other methods in both metrics, which proves it generates shapes with higher fidelity and better diversity.
}
\label{tab:main_com}
\begin{center}
{
\begin{tabular}{c|ccccccccc}
\toprule
Method & CF-\textit{MS} & CF-\textit{G} & CF-\textit{TG} & CF-\textit{CG} & ZS-ASDF &CS-\textit{const.} & CS-\textit{sqrt} & CS-\textit{linear} & CS-\textit{cosine}\\
\midrule

\textbf{FID} $\downarrow$ & 2425.25 & 2233.48 & 2141.61 & 2100.67 & 7332.93 & 1821.78 & \textbf{1480.11} & 1629.51 & 1725.63 \\
\textbf{ACC} $\uparrow$ & 83.33 & 62.81 & 68.71 & 71.11 & 39.14 & 86.59 & 87.08 & \textbf{87.50} & 87.27 \\

\bottomrule
\end{tabular}

}
\end{center}
\end{table*}

\noindent \textbf{Baseline. }
We compare the performance of our method against CLIP-Forge ~\cite{sanghi2022clip}, CLIP-Mesh ~\cite{khalid2022clip}
and DreamFields \cite{jain2022zero},
which are currently state of the art for zero-shot text-to-shape generation. We also set up a comparison method called Zero-Shot AutoSDF(ZS-ASDF), which trains the supervised method AutoSDF\cite{mittal2022autosdf} with the CLIP features as \nickname, to highlight the superiority of our conditional generation model. CLIP-Forge reports results on single shape generation for 234 text queries. These single shapes were generated using the mean of the prior (the Gaussian distribution), so we refer to these results as CF-MS. Note that this does not capture the diversity of shapes that can be generated given a text query. To capture results on more shape generations, we sample 32 shapes instead of one using a Gaussian (CF-G), truncated Gaussian (CF-TG) or clipped Gaussian (CF-CG) distribution. We also generate 32 shapes for ZS-ASDF and our method for fair comparison. We only compare qualitatively with DreamFields and CLIP-Mesh as the optimization time per query is significant (Table \ref{tab:intro_table}), and it does not use prior knowledge from the shape dataset.

\subsection{Evaluating Shape Diversity and Accuracy}

In this section, we first quantitatively evaluate the diversity and accuracy of a given shape matching a given text query on the ShapeNet13 dataset. We compare with CLIP-Forge and ZS-ASDF using the \textbf{Acc} and \textbf{FID} metrics. The results are shown in Table~\ref{tab:main_com}.  The first four columns represent CLIP-Forge(CF) and its different sampling techniques. The fifth column represents the result from AutoSDF trained with CLIP features(ZS-ASDF).  The other columns represent our method with different annealing scale strategies. Three major things can be observed. First, all variants of our method outperform CLIP-Forge and ZS-ASDF significantly. This indicates that the method produces more diverse and higher fidelity shapes of increasing accuracy. Second, it can be seen that annealing strategies for guidance scale give a better accuracy-diversity trade-off than constant scale guidance. Finally, we note that simply augmenting AutoSDF (another VQ-VAE) with CLIP features does not lead to improved text to shape generation, indicating the importance of our architecture design decisions. 

We also qualitatively compare our method to CLIP-Forge, CLIP-Mesh, and DreamFields. The results are shown in Figure \ref{fig:qual_multiple}. It can be observed that our method generates higher quality shapes (view ``a machine gun"), with more detail (view ``an office chair") and higher diversity (view ``a jet"). CLIP-Forge usually produces the same shape with small variations (view ``a rectangular table"). We also note that DreamFields and CLIP-Mesh produce very abstract results which may not be useful in many applications. Finally, we also show results on ShapeNet55 in the last row of Figure \ref{fig:qual_multiple}. We could not get CLIP-Forge to produce sensible shapes for most text queries on ShapeNet55 (which we attribute to the data imbalance issue of ShapeNet55), whereas our method produced high fidelity shapes.

\setlength{\tabcolsep}{1pt}
\begin{table}
\caption{Ablations on the major components of \nickname. \nickname achieves the best performance comparing with other design choice, which proves the effectiveness of the major components.}

\begin{subtable}[c]{.5\textwidth}
\label{tab:layer00}
\subcaption{Effect of varying the \textit{Noise} parameter.}
\centering
\begin{tabular}{c|cccccc}
\toprule
\textit{$\gamma$} & $\times$ & 0.5 & 0.8 & 1.0 & 1.2 & 1.5 \\
\midrule
\textbf{FID$\downarrow$} & 1720.02 & 1764.98 & 1484.61 & 1703.38 & \textbf{1447.91} & 1478.17\\
\textbf{ACC$\uparrow$} & 64.87 & 75.73 & 77.41 & 79.09 & \textbf{79.63} & 78.47\\
\bottomrule
\end{tabular}

\end{subtable}

\begin{subtable}[c]{.5\textwidth}
\label{tab:layer01}
\subcaption{Effect of varying number of layers  (\textit{L}) in the mapping network.}
\centering
\begin{tabular}{c|cccccc}
\toprule
\textit{L} & 0 & 1 & 2 & 3 & 4 & 5  \\
\midrule
\textbf{FID$\downarrow$} & 2874.87 & 1716.73 & 1518.97 & 1447.91 & 1532.50 & \textbf{1424.46}\\
\textbf{ACC$\uparrow$} & 62.72 & 78.70 & 79.17 & \textbf{79.63} & 77.16 & 76.09\\
\bottomrule
\end{tabular}
\end{subtable}

\setlength{\tabcolsep}{9.4pt}
\begin{subtable}[c]{.5\textwidth}
\label{tab:layer02}
\subcaption{Comparison  with baselines on super resolution.}
\centering
\begin{tabular}{c|ccc}
\toprule
\textit{Method} & 3D-UNet & $64^3$ -DS & \nickname \\
\midrule
\textbf{FID$\downarrow$} & 2056.92 & 2196.96 & \textbf{1910.28} \\
\textbf{ACC$\uparrow$} & 86.65 & 77.92 & \textbf{86.85} \\
\bottomrule
\end{tabular}

\end{subtable}

\label{tab:layers}
\end{table}
\subsection{Major Components for Conditional Generation}

\noindent \textbf{Effect of Noise Parameter. }
During Stage 2 Training, to better align the image and text embeddings, we added Gaussian noise at varying levels of $\gamma$. We investigate the effect of ($\gamma$) added to the image condition vector and show the results in Table \ref{tab:layers} (a). We keep the number of mapping layers fixed at three in this experiment. We observe that adding Gaussian noise drastically improves both the diversity and accuracy of generation, with the optimal noise parameter being around 1.2. These results indicate that adding noise during training helps align the text features seen at inference and the image features seen at training.

\noindent \textbf{Size of Mapping Network.}
We next probe the importance of the mapping network. We show the results in Table \ref{tab:layers} (b), where \textit{L} represents the number of layers of the  mapping function. In the case of 0 layers, we directly project conditional embeddings to the layernorm parameters using a linear layer. We make two observations from the results: 1) A common mapping network improves both the accuracy and FID. 2) Increasing the number of mapping network layers beyond a certain number decreases accuracy. 

\noindent \textbf{Classifier-Free Guidance.}
We next explore the relationship between dropping out image conditioning at $\rho\%$ during training and using \textit{constant} classifier-free guidance during test time. In Table \ref{tab:classifier_free}, we vary the scale parameter, $a(t)=k$, across the columns. It can be seen that as we increase the scale, accuracy typically increases  while FID decreases. This indicates that the method is giving more accurate results on a given text query at the cost of shape diversity. A low dropout rate (5-15\%) gives a good trade-off between accuracy and diversity. 

\begin{table*}
\setlength{\tabcolsep}{10pt}
\caption{Classifier-Free Guidance and Step-Unrolled Training (SUT) experiment results. $\rho\%$ is the drop out rate for classifier-free guidance. SUT indicates the use of Step-Unrolled Training. The remaining columns indicate the variation in scale parameter. }
\label{tab:classifier_free}
\centering
{\small
  \begin{tabular}{l|l|ll|ll|ll|ll|ll}
    \toprule
    \multirow{2}{*}{ $\rho \%$ } &
    \multirow{2}{*}{SUT} &
      \multicolumn{2}{c|}{$a(t)=3$} &
      \multicolumn{2}{c|}{$a(t)=2.5$} &
      \multicolumn{2}{c|}{$a(t)=2$}  &
      \multicolumn{2}{c|}{$a(t)=1.5$}  &
      \multicolumn{2}{c}{$a(t)=1$}  
      \\
      
    &  & \textbf{FID$\downarrow$} & \textbf{Acc$\uparrow$} & \textbf{FID$\downarrow$} & \textbf{Acc$\uparrow$} & \textbf{FID$\downarrow$} & \textbf{Acc$\uparrow$} & \textbf{FID$\downarrow$} & \textbf{Acc$\uparrow$} & \textbf{FID$\downarrow$} & \textbf{Acc$\uparrow$} \\
    \midrule
    5 & $\times$ & 2059.0 & 85.73 & 1970.1 & 85.47 & 1790.9 & 85.43 & 1536.5 &	83.98 & 1227.5	& 79.17\\
     10 & $\times$ & 1893.2 & 84.46 & 1821.2 & 84.50 & 1684.3 & 83.41 & 1522.4	 & 82.34  & 1348.4 &	77.76\\
      15 & $\times$ & 2086.9 &	85.03 & 1964.7 &	84.99 & 1851.9 &	84.36 & 1660.9 &	83.14  & 1485.5 &	78.19\\
      20 & $\times$ & 2062.5 &	83.39 & 1972.3	& 82.95 & 1892.9 & 82.74 & 1733.3 &	81.13 & 1566.6 & 75.94\\
      5 & $\checkmark$ & 2039.8 & 87.69 & 2011.1 &	87.39 & 1811.8 & 87.40 & 1678.6	 & 86.24 & 1517.9 & 82.18\\      
    \hline
  \end{tabular}
  }
\end{table*}

\begin{figure*}
\centering
\includegraphics[width=\textwidth]{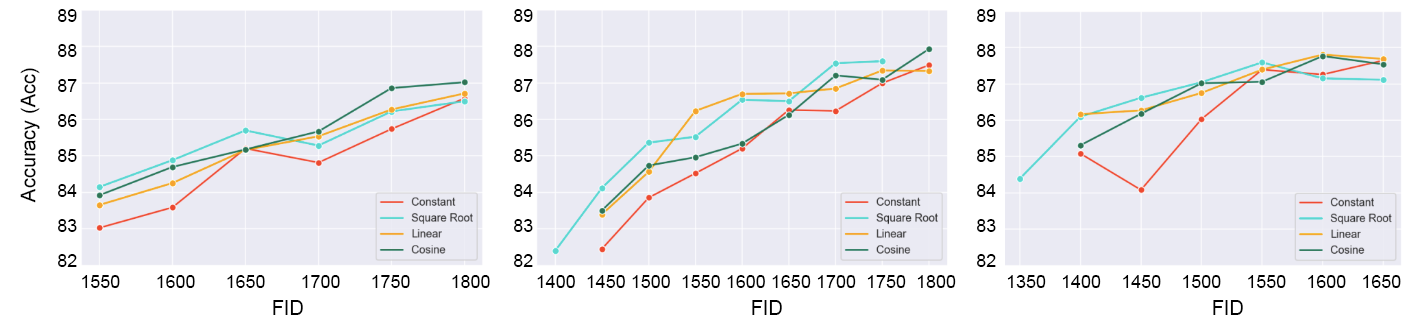}
\caption{\textbf{FID} and \textbf{Acc} results with different classifier-free guidance scale annealing strategies (constant, cosine, square root, linear) across three different runs (different seeds) of the \nickname~Stage 2 Coarse Transformer.
}
\label{fig:three-graphs}
\end{figure*}

\noindent \textbf{Step-Unrolled Training (SUT).}
 Finally, to further improve the accuracy we investigate the use of step-unrolled training~\cite{step_unrolled}. 
The results are shown in the last row of Table \ref{tab:classifier_free}. From the table it can be observed that step-unrolled training enables higher accuracy across all scales. This indicates that the model learns to unmask samples it would encounter during inference time.

\subsection{Annealing Strategy for Scale Parameter}
An important idea we propose in this paper is a scale annealing technique for classifier-free guidance. We employ classifier-free guidance to identify a better accuracy-diversity trade-off with different annealing schedules: constant, linear, cosine and square root. 
To determine which annealing technique works the best, we fix the FID values and plotted the accuracy at each fixed FID value (Figure \ref{fig:three-graphs}). 
As different scale parameters give different FID values, we conducted an extensive grid search over the starting scale parameter.
Note that finding the exact FID is not always feasible, so we pick the closest FID. Figure \ref{fig:three-graphs} shows results of FID versus accuracy on three different runs of the Coarse Transformer.  We find that across all three runs, the accuracy is typically lower for a given FID for constant schedules as opposed to other schedules. This is especially the case at the lower range of FID values. The results indicate that having a large scale at the beginning of sampling  is more important than later stages, especially in use cases where diversity is paramount.

\subsection{Super-Resolution }

Finally, we investigate the importance of hierarchical super-resolution. We compare our method with two baselines. In the first baseline, we directly use the $32^3$ resolution results from the coarse transformer and use a 3D-UNET  based super-resolution network to translate from $32^3$ to $64^3$.  For the second baseline, we train a transformer directly on $64^3$ VQ-VAE that is conditioned on text features as opposed to a coarse resolution  grid. We refer to this as $64^3$-DS ($64^3$ direct synthesis).  The results are shown in Table \ref{tab:layers}(c). It can be seen from the table that latent-based super-resolution outperforms the baselines in both accuracy and diversity.

\section{Conclusion}
We present \nickname, a text-to-3D-shape generation method capable of producing shapes of high fidelity and diversity without the need for (text, shape) pairs.
To achieve this, \nickname leverages CLIP and implements super-resolution in a discrete latent space with a hierarchical architecture and a novel annealed variant of classifier-free guidance on a mask-based model.
We validate \nickname by comparing it with a number of baselines in terms of FID and accuracy, finding that \nickname is the new state of the art for this problem.
In experimenting with different guidance scale scheduling, we find that constant scale scheduling did not always work the best, an important finding for the diffusion modeling community that may improve generation quality.
Our paper is an important step towards improved quality and diversity of 3D text-to-shape generation outcomes.

\noindent \paragraph{Limitations and Societal Impact.} Despite providing higher quality over baseline methods, \nickname still suffers from limitations in its ability to capture smaller details (e.g.,~chair with a hole on its back) and generate shapes from text prompts involving counting (e.g.,~chair with \emph{four} slats). Furthermore, \nickname also fails on many text queries which are not present in the prior knowledge of CLIP. The method also lacks the ability to produce texture. In future work, we will address these limitations by exploring implicit representations that can capture even finer details and by investigating neural networks that can count. 

Text-to-shape approaches can make the generation of 3D content more accessible and efficient. Within the text-to-image domain there has been concern that automatic approaches may replace artists. Because our approach produces voxels, a representation that lends well to direct manipulation and handoff to 3D designers, we believe that our text-to-3D approach can augment 3D design workflows rather than replace them.

\paragraph{Acknowledgements}
This research was supported by AFOSR grant FA9550-21-1-0214.

{\small
\bibliographystyle{ieee_fullname}
\bibliography{egbib}
}

\end{document}